\documentclass[sigconf]{acmart}
\usepackage{balance}
\def\BibTeX{{\rm B\kern-.05em{\sc i\kern-.025em b}\kern-.08emT\kern-.1667em\lower.7ex\hbox{E}\kern-.125emX}}

\usepackage{amsmath,amssymb,amsfonts}
\usepackage{algorithmic}
\usepackage{graphicx}
\usepackage{textcomp}
\usepackage{amsmath}
\usepackage{bbm}
\usepackage{subcaption}
\usepackage{url}
\usepackage{multirow}
\usepackage{amssymb}

\newcommand{\bfx}{\mathbf{x}}
\newcommand{\bfz}{\mathbf{z}}
\DeclareMathOperator*{\argmax}{argmax}

\setcopyright{rightsretained}
\begin{document}

\fancyhead{}
\title{Privacy Risks of Securing Machine Learning Models against Adversarial Examples}
\author{Liwei Song}
\email{liweis@princeton.edu}
\affiliation{Princeton University}
\author{Reza Shokri}
\email{reza@comp.nus.edu.sg}
\affiliation{National University of Singapore}
\author{Prateek Mittal}
\email{pmittal@princeton.edu}
\affiliation{Princeton University}

\copyrightyear{2019} 
\acmYear{2019} 
\acmConference[CCS '19]{2019 ACM SIGSAC Conference on Computer and Communications Security}{November 11--15, 2019}{London, United Kingdom}
\acmBooktitle{2019 ACM SIGSAC Conference on Computer and Communications Security (CCS '19), November 11--15, 2019, London, United Kingdom}\acmDOI{10.1145/3319535.3354211}
\acmISBN{978-1-4503-6747-9/19/11}

\begin{abstract}
The arms race between attacks and defenses for machine learning models has come to a forefront in recent years, in both the security community and the privacy community.
However, one big limitation of previous research is that the security domain and the privacy domain have typically been considered \emph{separately}. 
It is thus unclear whether the defense methods in one domain will have any unexpected impact on the other domain.

In this paper, we take a step towards resolving this limitation by combining the two domains. 
In particular, we measure the success of \emph{membership inference attacks} against six state-of-the-art \emph{defense methods 
that mitigate the risk of adversarial examples (i.e., evasion attacks)}.
Membership inference attacks determine whether or not an individual data record has been part of a model's training set. The accuracy of such attacks reflects the information leakage of training algorithms about individual members of the training set.
Adversarial defense methods against adversarial examples influence the model's decision boundaries such that model predictions remain unchanged 
for a small area around each input. However, this objective is optimized on training data.  Thus, individual data records in the training set have a significant influence on robust models. This makes the models more vulnerable to inference attacks.

To perform the membership inference attacks, we leverage the existing inference methods that exploit model predictions. We also propose 
two new inference methods that exploit \emph{structural properties of robust models on adversarially perturbed data}. 
Our experimental evaluation demonstrates that compared with the natural training (undefended) approach, 
\emph{adversarial defense methods can indeed increase the target model's risk against membership inference attacks}.
When using adversarial defenses to train the robust models, the membership inference advantage increases by up to $4.5$ times compared to the naturally undefended models. 
Beyond revealing the privacy risks of adversarial defenses, we further investigate the factors, such as model capacity, that influence the membership information leakage.

\end{abstract}

\ccsdesc[500]{Security and privacy~Software and application security}
\ccsdesc[500]{Computing methodologies~Neural networks}
\keywords{machine learning; membership inference attacks; adversarial examples and defenses}

\maketitle

\section{Introduction}

Machine learning models, especially deep neural networks, have been deployed prominently in many real-world applications, 
such as image classification \cite{krizhevsky_imagenet_NIPS12, simonyan_DNN_arxiv14}, 
speech recognition \cite{hinton_NNspeech_magazine12, deng_NNspeech_ICASSP13}, 
natural language processing \cite{collobert_NLP_ML11, andor_NLP_arxiv16}, and game playing \cite{silver_alphago_nature16, moravvcik_poker_AAAS17}.
However, since the machine learning algorithms were originally designed without considering potential adversarial threats, 
their security and privacy vulnerabilities have come to a forefront in recent years, 
together with the arms race between attacks and defenses \cite{huang_advML_AIsec11, biggio_AML_2018, papernot_SOK_EuroSP18}.

In the security domain, the adversary aims to induce misclassifications to the target machine learning model, 
with attack methods divided into two categories: evasion attacks and poisoning attacks \cite{huang_advML_AIsec11}.
Evasion attacks, also known as adversarial examples, perturb inputs at the test time to induce wrong predictions by the target model 
\cite{biggio_evasion_KDD13, szegedy_evasion_arxiv13, carlini_evasion_SP17, Goodfellow_evasion_arxiv14, papernot_adv_EuroSP16}.
In contrast, poisoning attacks target the training process by maliciously modifying part of training data to cause the trained model 
to misbehave on some test inputs \cite{biggio_poison_ICML12, koh_poison_ICML17, shafahi_poison_NIPS18}.
In response to these attacks, the security community has designed new training algorithms to secure machine learning models 
against evasion attacks \cite{madry_robust_ICLR18, sinha_PAT_ICLR18, zhang_TRADES_ICML19, wong_robust_ICML18, mirman_provable_ICML18, gowal_IBP_secml18} 
or poisoning attacks \cite{steinhardt_defense_poison_NIPS17, jagielski_defense_poison_SP18}.

In the privacy domain, the adversary aims to obtain private information about the model's training data or the target model.
Attacks targeting data privacy include: 
the adversary inferring whether input examples were used to train the target model with membership inference attacks 
\cite{shokri_membership_SP17, yeom_privacy_CSF18, nasr_whitebox_privacy_arxiv18}, 
learning global properties of training data with property inference attacks \cite{ganju_property_privacy_CCS18}, or covert channel model training attacks \cite{song_active_privacy_CCS17}.
Attacks targeting model privacy include: 
the adversary uncovering the model details with model extraction attacks \cite{tramer_extraction_USENIX16}, 
and inferring hyperparameters with hyperparameter stealing attacks \cite{wang_steal_SP18}.
In response to these attacks, the privacy community has designed defenses to prevent privacy leakage 
of training data \cite{nasr_membership_defense_CCS18, hayes_membership_defense_NIPS18, shokri_privacy_dl_CCS15, abadi_DP_DL_CCS16} 
or the target model \cite{kesarwani_extraction_warn_ACSAC18, lee_steal_defense_DLS19}.

\begin{figure*}[!ht]
	\centering
	\begin{subfigure}[t]{0.48\linewidth}
		\raggedleft
		\includegraphics[width=\linewidth]{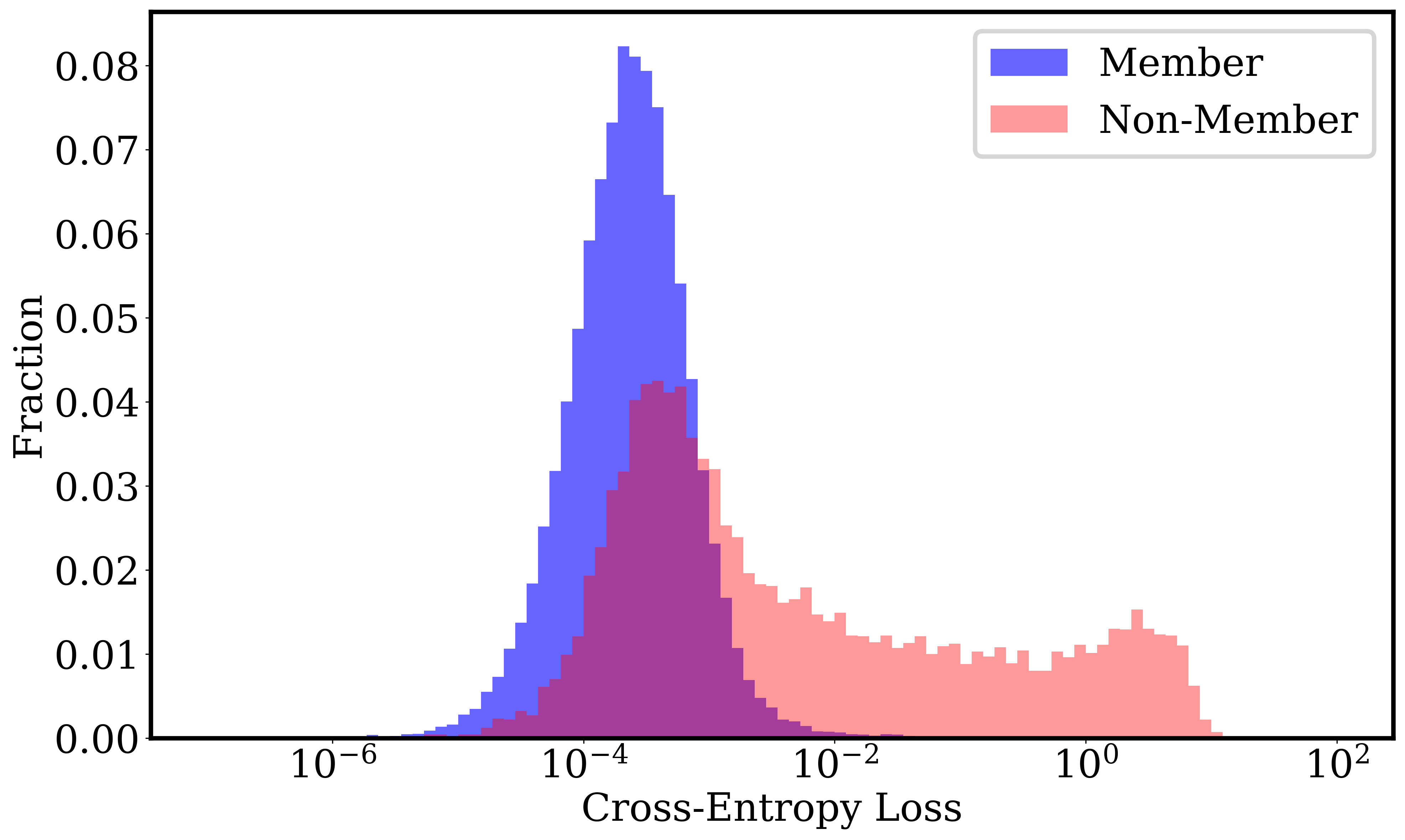}
		\caption{Adversarially robust model from Madry et al. \cite{madry_robust_ICLR18}, with $99\%$ train accuracy and $87\%$ test accuracy.}
		\label{fig:loss_adv_cifar}
	\end{subfigure}\hfill
	\begin{subfigure}[t]{0.48\linewidth}
		\raggedright
		\includegraphics[width=\linewidth]{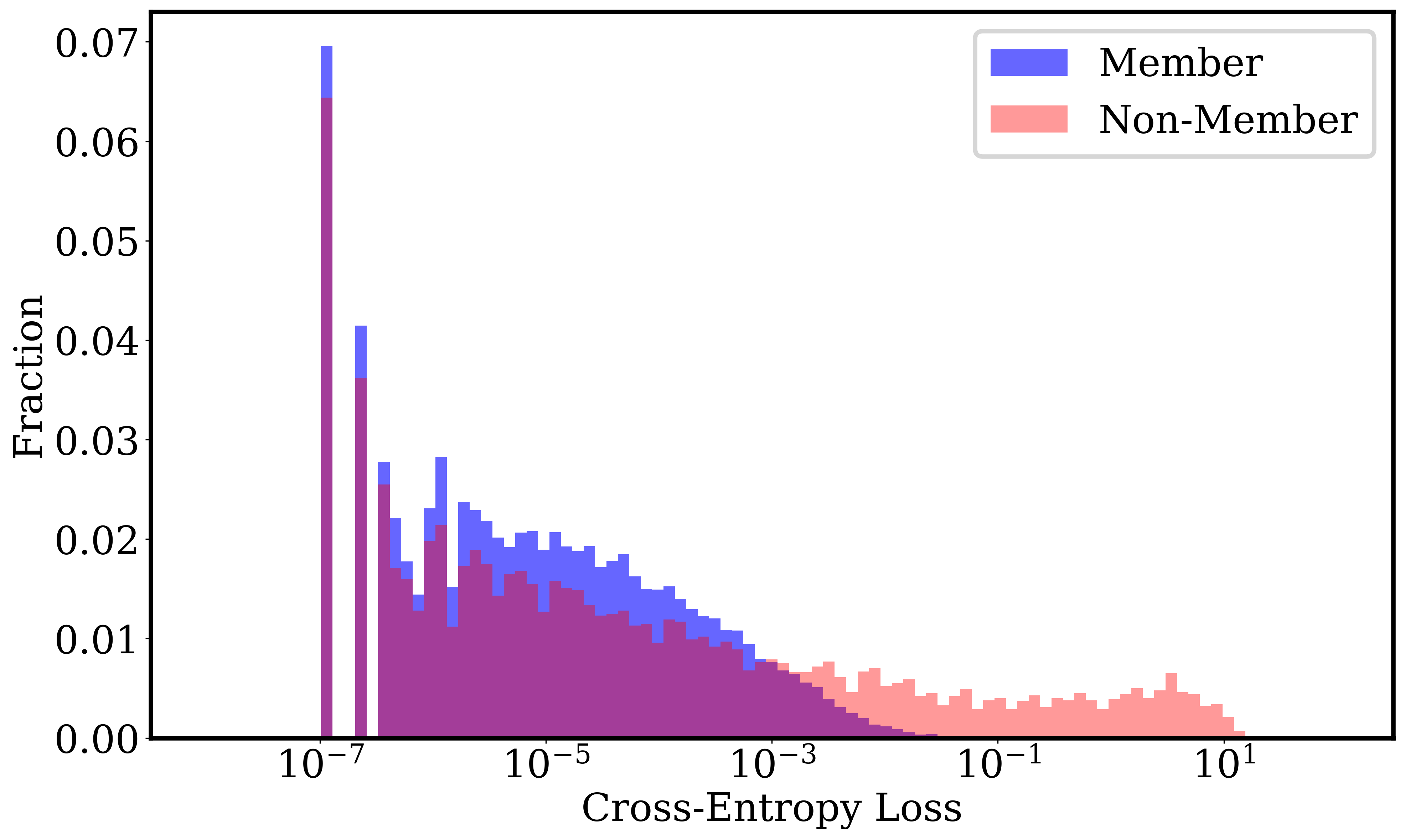}
		\caption{Naturally undefended model, with $100\%$ train accuracy and $95\%$ test accuracy. Around $23\%$ training and test examples have zero loss.}
		\label{fig:loss_nat_cifar}
	\end{subfigure}
	\caption{Histogram of CIFAR10 classifiers' loss values of training data (members) and test data (non-members). We can see the larger divergence between the loss distribution over members and non-members on the robust model as compared to the natural model. This shows the privacy risk of securing deep learning models against adversarial examples.
	}
	\label{fig:loss_for_privacy}
\end{figure*}

However, one important limitation of current machine learning defenses is that they typically focus solely on either the security domain or the privacy domain. 
It is thus unclear whether defense methods in one domain will have any unexpected impact on the other domain.
In this paper, we take a step towards enhancing our understanding of machine learning models when both the security domain and privacy domain are considered together.
In particular, we seek to understand the privacy risks of securing machine learning models by evaluating 
\emph{membership inference attacks against adversarially robust deep learning models}, which aim to mitigate the threat of adversarial examples.

The membership inference attack aims to infer whether a data point is part of the target model's training set or not, reflecting the information leakage of the model about its training data. It can also pose a privacy risk as the membership can reveal an individual's sensitive information. 
For example,  participation in a hospital's health analytic training set means that an individual was once a patient in that hospital.
It has been shown that the success of membership inference attacks in the black-box setting is highly related to the target model's generalization error 
\cite{shokri_membership_SP17, yeom_privacy_CSF18}. 
Adversarially robust models aim to enhance the robustness of target models by ensuring that model predictions are unchanged for 
a small area (such as $l_{\infty}$ ball) around each input example. The objective is to make the model robust against any input, however, the objective is optimized only on the training set. Thus, intuitively, adversarially robust models have the potential to increase the model's generalization error and sensitivity to changes in the training set, 
resulting in an enhanced risk of membership inference attacks.
As an example, Figure~\ref{fig:loss_for_privacy} shows the histogram of cross-entropy loss values of training data and test data for 
both naturally undefended and adversarially robust CIFAR10 classifiers provided by Madry et al. \cite{madry_robust_ICLR18}. 
We can see that members (training data) and non-members (test data) can be distinguished more easily for the robust model, compared to the natural model.

To measure the membership inference risks of adversarially robust models, besides the conventional inference method based on prediction confidence, we propose two new inference methods that exploit the structural properties of robust models. 
We measure the privacy risks of robust models trained with six state-of-the-art adversarial defense methods, 
and find that adversarially robust models are indeed more susceptible to membership inference attacks than naturally undefended models.
We further perform a comprehensive investigation to analyze the relation between privacy leakage and model properties.
We finally discuss the role of adversary's prior knowledge, potential countermeasures and the relationship between privacy and robustness.

In summary, we make the following contributions in this paper:
\begin{enumerate}
\item {We propose two new membership inference attacks specific to adversarially robust models by exploiting adversarial examples' predictions and verified worst-case predictions. 
With these two new methods, we can achieve higher inference accuracies than the conventional inference method based on prediction confidence of benign inputs.}
\item {We perform membership inference attacks on models trained with six state-of-the-art adversarial defense methods 
(3 empirical defenses \cite{madry_robust_ICLR18, sinha_PAT_ICLR18, zhang_TRADES_ICML19} and 
3 verifiable defenses \cite{wong_robust_ICML18, mirman_provable_ICML18, gowal_IBP_secml18}). 
We demonstrate that all methods indeed increase the model's membership inference risk.
By defining the membership inference advantage as the increase in inference accuracy over random guessing (multiplied by 2) \cite{yeom_privacy_CSF18}, 
we show that robust machine learning models can incur a membership inference advantage $4.5 \times$, $2\times$, $3.5 \times$ times 
the membership inference advantage of naturally undefended models, on Yale Face, Fashion-MNIST, and CIFAR10 datasets, respectively.}
\item {We further explore the factors that influence the membership inference performance of the adversarially robust model, 
including its robustness generalization, the adversarial perturbation constraint, and the model capacity.
}
\item {Finally, we experimentally evaluate the effect of the adversary's prior knowledge, countermeasures such as temperature scaling and regularization, and discuss the relationship between training data privacy and model robustness.
} 
\end{enumerate}

Some of our analysis was briefly discussed in a short workshop paper \cite{anony_security_vs_privacy_19}.
In this paper, we go further by proposing two new membership inference attacks and measuring four more adversarial defense methods, 
where we show that all adversarial defenses can increase privacy risks of target models.
We also perform a comprehensive investigation of factors that impact the privacy risks.

\section{Background and Related Work: Adversarial Examples and Membership Inference Attacks}

In this section, we first present the background and related work on adversarial examples and defenses, and then discuss membership inference attacks.

\subsection{Adversarial Examples and Defenses}\label{subsec:adv_attack}

Let $F_{\theta}: \mathbbm{R}^{d} \rightarrow \mathbbm{R}^{k}$ be a machine learning model with $d$ input features and $k$ output classes, 
parameterized by weights $\theta$.
For an example $\bfz = (\bfx, y)$ with the input feature $\bfx$ and the ground truth label $y$, 
the model outputs a prediction vector over all class labels $F_{\theta}(\bfx)$ with $\sum_{i=0}^{k-1} F_{\theta}(\bfx)_{i} = 1$, 
and the final prediction will be the label with the largest prediction probability $\hat{y} = \argmax_{i} F_{\theta}(\bfx)_{i}$.
For neural networks, the outputs of its penultimate layer are known as logits, and we represent them as a vector $g_{\theta}(\bfx)$. 
The softmax function is then computed on logits to obtain the final prediction vector.
\begin{equation}\label{eq:softmax}
F_{\theta}(\bfx)_{i} = \frac{\exp \,(g_{\theta}(\bfx)_{i})}{\sum_{j=0}^{k-1} \exp \,(g_{\theta}(\bfx)_{j})}
\end{equation}

Given a training set $D_{\text{train}}$, the natural training algorithm aims to make model predictions match ground truth labels 
by minimizing the prediction loss over all training examples.
\begin{equation}\label{eq:nat-train}
\min_{\theta} \, \frac{1}{|D_{\text{train}}|}\sum_{\bfz \in D_{\text{train}}} \mathcal{\ell}(F_{\theta},\bfz),
\end{equation}
where $|D_{\text{train}}|$ denotes the size of training set, and $\ell$ computes the prediction loss. A widely-adopted loss function is the cross-entropy loss:
\begin{equation}\label{eq:cross_entropy_loss}
\mathcal{\ell}(F_{\theta},\bfz) = -\sum_{i=0}^{k-1} \mathbbm{1}\{i=y\} \cdot \log(F_{\theta}(\bfx)_{i}),
\end{equation}
where $\mathbbm{1\{\cdot\}}$ is the indicator function.

\subsubsection{Adversarial examples:} 
Although machine learning models have achieved tremendous success in many classification scenarios, they have been found to be easily fooled 
by adversarial examples \cite{szegedy_evasion_arxiv13, biggio_evasion_KDD13, Goodfellow_evasion_arxiv14, carlini_evasion_SP17, papernot_adv_EuroSP16}. 
Adversarial examples induce incorrect classifications to target models, and can be generated via imperceptible perturbations to benign inputs. 
\begin{equation}\label{eq:adv-example}
\argmax_{i} \, F_{\theta}(\tilde{\bfx})_{i} \neq y, \quad \text{such that} \, \, \tilde{\bfx} \in \mathcal{B}_{\epsilon}(\bfx),
\end{equation}
where $\mathcal{B}_{\epsilon}(\bfx)$ denotes the set of points around $\bfx$ within the perturbation budget of $\epsilon$. 
Usually a $l_{p}$ ball is chosen as the perturbation constraint for generating adversarial examples
i.e., $\mathcal{B}_{\epsilon}(\bfx) = \{\bfx' \, | \, \lVert \bfx'- \bfx \rVert_{p} \leq \epsilon \}$. 
We consider the $l_{\infty}$-ball adversarial constraint throughout the paper, as it is widely adopted 
by most adversarial defense methods \cite{madry_robust_ICLR18, sinha_PAT_ICLR18, zhang_TRADES_ICML19, wong_robust_ICML18, mirman_provable_ICML18, gowal_IBP_secml18, raghunathan_certified_ICLR18}.

The solution to Equation \eqref{eq:adv-example} is called an ``untargeted adversarial example'' as the adversarial goal is to achieve any incorrect classification. 
In comparison, a ``targeted adversarial example'' ensures that the model prediction is a specified incorrect label $y'$, which is not equal to $y$.
\begin{equation}\label{eq:tar-adv-example}
\argmax_{i} \, F_{\theta}(\tilde{\bfx})_{i} = y', \quad \text{such that} \, \, \tilde{\bfx} \in \mathcal{B}_{\epsilon}(\bfx).
\end{equation}
Unless otherwise specified, an adversarial example in this paper refers to an untargeted adversarial example.

To provide adversarial robustness under the perturbation constraint $\mathcal{B}_{\epsilon}$, 
instead of natural training algorithm shown in Equation \eqref{eq:nat-train}, a robust training algorithm is adopted by adding an additional robust loss function.
\begin{equation}\label{eq:rob-train}
\min_{\theta} \, \frac{1}{|D_{\text{train}}|} \sum_{\bfz \in D_{\text{train}}} \alpha \cdot \ell (F_{\theta},\bfz) + (1-\alpha) \cdot \ell_{R}(F_{\theta},\bfz, \mathcal{B}_{\epsilon}),
\end{equation}
where $\alpha$ is the ratio to trade off natural loss and robust loss, and $\ell_{R}$ measures the robust loss, 
which can be formulated as maximizing prediction loss $\ell'$ under the constraint $\mathcal{B}_{\epsilon}$.
\begin{equation}\label{eq:exact_robust_loss}
\ell_{R}(F_{\theta},\bfz, \mathcal{B}_{\epsilon}) = \max_{\tilde{\bfx} \in \mathcal{B}_{\epsilon}(x)} \ell' (F_{\theta},(\tilde{\bfx}, y))
\end{equation}
$\ell'$ can be same as $\ell$ or other appropriate loss functions.

However, it is usually hard to find the exact solution to Equation \eqref{eq:exact_robust_loss}. 
Therefore, the adversarial defenses propose different ways to approximate the robust loss $\ell_{R}$, 
which can be divided into two categories: empirical defenses and verifiable defenses.

\subsubsection{Empirical defenses:}\label{subsub:empirical_defense}

Empirical defense methods approximate robust loss values by generating adversarial examples $\bfx_{adv}$ at each training step 
with state-of-the-art attack methods and computing their prediction loss. Now the robust training algorithm can be expressed as following.
\begin{equation}
\min_{\theta} \frac{1}{|D_{\text{train}}|} \sum_{\bfz \in D_{\text{train}}} \alpha \cdot \ell (F_{\theta},\bfz) + (1-\alpha) \cdot \ell'(F_{\theta},(\bfx_{adv}, y))
\end{equation}

Three of our tested adversarial defense methods belong to this category, which are described as follows.

\noindent \textbf{PGD-Based Adversarial Training (PGD-Based Adv-Train)} \cite{madry_robust_ICLR18}: 
Madry et al. \cite{madry_robust_ICLR18} propose one of the most effective empirical defense methods by using the projected gradient descent (PGD) method 
to generate adversarial examples for maximizing cross-entropy loss ($\ell'=\ell$) and training purely on those adversarial examples ($\alpha=0$). 
The PGD attack contains $T$ gradient descent steps, which can be expressed as
\begin{equation}\label{eq:PGD}
\tilde{\bfx}^{t+1} = \Pi_{\mathcal{B}_{\epsilon}(\bfx)} [\tilde{\bfx}^{t} + \eta \cdot \text{sign}(\, \nabla_{\tilde{\bfx}^{t}} \ell(F_{\theta},(\tilde{\bfx}^{t}, y) ) \,) ],
\end{equation}
where $\tilde{\bfx}^{0} = \bfx$, $\bfx_{adv} = \tilde{\bfx}^{T}$, $\eta$ is the step size value, $\nabla$ denotes the gradient computation, 
and $\Pi_{\mathcal{B}_{\epsilon}(\bfx)}$ means the projection onto the perturbation constraint $\mathcal{B}_{\epsilon}(\bfx)$.

\noindent \textbf{Distributional Adversarial Training (Dist-Based Adv-Train)} \cite{sinha_PAT_ICLR18}: 
Instead of strictly satisfying the perturbation constraint with projection step $\Pi_{\mathcal{B}_{\epsilon}(\bfx)}$ as in PGD attacks, 
Sinha et al. \cite{sinha_PAT_ICLR18} generate adversarial examples by solving the Lagrangian relaxation of cross-entropy loss:
\begin{equation}\label{eq:wrm-attack}
\max_{\tilde{\bfx}} \, \ell(F_{\theta}, (\tilde{\bfx}, y)) - \gamma \lVert \tilde{\bfx}-\bfx \rVert_{p},
\end{equation}
where $\gamma$ is the penalty parameter for the $l_{p}$ distance. A multi-step gradient descent method is adopted to solve Equation \eqref{eq:wrm-attack}. 
The model will then be trained on the cross-loss entropy ($\ell' = \ell$) of adversarial examples only ($\alpha=0$).

Sinha et al. \cite{sinha_PAT_ICLR18} derive a statistical guarantee for $l_{2}$ distributional robustness with strict conditions 
requiring the loss function $\ell$ to be smooth on $\bfx$, which are not satisfied in our setting. 
We mainly use widely-adopted ReLU activation functions for our machine learning models, which result in a non-smooth loss function. 
Also, we generate adversarial examples with $l_{\infty}$ distance penalties by using the algorithm proposed by Sinha et al. \cite{sinha_PAT_ICLR18} in Appendix E, 
where there is no robustness guarantee. Thus, we categorize the defense method as empirical.

\noindent \textbf{Difference-based Adversarial Training (Diff-Based Adv-Train)} \cite{zhang_TRADES_ICML19}:
Instead of using the cross-entropy loss of adversarial examples, with insights from a toy binary classification task, 
Zhang et al. \cite{zhang_TRADES_ICML19} propose to use the difference (e.g., Kullback-Leibler (KL) divergence) 
between the benign output $F_{\theta}(\bfx)$ and the adversarial output $F_{\theta}(\bfx_{adv})$ as the loss function $\ell'$, 
and combine it with natural cross entropy loss ($\alpha \neq 0$).
\begin{equation}
\ell'(F_{\theta},(\bfx_{adv}, y)) = d_{kl}(F_{\theta}(\bfx_{adv}), F_{\theta}(\bfx)),
\end{equation}
where $d_{kl}$ computes the KL divergence. Adversarial examples are also generated with PGD-based attacks, 
except that now the attack goal is to maximize the output difference, 
\begin{equation}\label{eq:Diff-PGD}
\tilde{\bfx}^{t+1} = \Pi_{\mathcal{B}_{\epsilon}(\bfx)} [\tilde{\bfx}^{t} + \eta \cdot \text{sign}(\, \nabla_{\tilde{\bfx}^{t}} d_{kl}(F_{\theta}(\tilde{\bfx}^{t}), F_{\theta}(\bfx)) \,) ].
\end{equation}

\subsubsection{Verifiable defenses:}\label{subsubsec:verifiable_defense}
Although empirical defense methods are effective against state-of-the-art adversarial examples \cite{athalye_adv_ICML18}, there is no \emph{guarantee} for such robustness.
To obtain a guarantee for robustness, verification approaches have been proposed to compute an upper bound of prediction loss $\ell'$ 
under the adversarial perturbation constraint $\mathcal{B}_{\epsilon}$. 
If the input can still be predicted correctly in the verified worst case, then it is certain that there is no misclassification existing under $\mathcal{B}_{\epsilon}$.

Thus, verifiable defense methods take the verification process into consideration during training by using the verified worst case prediction loss as robust loss value $\ell_{R}$.
Now the robust training algorithm becomes
\begin{equation}
\min_{\theta} \frac{1}{|D_{\text{train}}|} \sum_{\bfz \in D_{\text{train}}} \alpha \cdot \ell (F_{\theta},\bfz) + (1-\alpha) \cdot \mathcal{V} (\ell' (F_{\theta},(\tilde{\bfx}, y) ), \mathcal{B}_{\epsilon} ),
\end{equation}
where $\mathcal{V}$ means verified upper bound computation of prediction loss $\ell'$ under the adversarial perturbation constraint $\mathcal{B}_{\epsilon}$. 
In this paper, we consider the following three verifiable defense methods.

\noindent \textbf{Duality-Based Verification (Dual-Based Verify)} \cite{wong_robust_ICML18}: 
Wong and Kolter \cite{wong_robust_ICML18} compute the verified worst-case loss by solving its dual problem with convex relaxation on non-convex ReLU operations 
and then minimize this overapproximated robust loss values only $(\alpha = 0, \ell' = \ell)$. 
They further combine this duality relaxation method with the random projection technique to scale to 
more complex neural network architectures \cite{wong_robust_NIPS18}, like ResNet \cite{he_ResNet_CVPR16}.

\noindent \textbf{Abstract Interpretation-Based Verification (Abs-Based Verify)} \cite{mirman_provable_ICML18}: 
Mirman et al. \cite{mirman_provable_ICML18} leverage the technique of abstract interpretation to compute the worse-case loss: 
an abstract domain (such as interval domain, zonotope domain \cite{gehr_ADP_SP18}) is used to express the 
adversarial perturbation constraint $\mathcal{B}_{\epsilon}$ at the input layer, and by applying abstract transformers on it, 
the maximum verified range of model output is obtained. 
They adopt a softplus function on the logits $g_{\theta}(\tilde{\bfx})$ to compute the robust loss value and then combine it with natural training loss ($\alpha \neq 0$).
\begin{equation}
\ell'(F_{\theta},(\tilde{\bfx}, y)) = \log \, ( \, \exp \, (\max_{y'\neq y} g_{\theta}(\tilde{\bfx})_{y'} - g_{\theta}(\tilde{\bfx})_{y}) + 1 \, )
\end{equation}

\noindent \textbf{Interval Bound Propagation-Based Verification (IBP-Based Verify)} \cite{gowal_IBP_secml18}: 
Gowal et al. \cite{gowal_IBP_secml18} share a similar design as Mirman et al. \cite{mirman_provable_ICML18}: 
they express the constraint $\mathcal{B}_{\epsilon}$ as a bounded interval domain (one specified domain considered by Mirman et al. \cite{mirman_provable_ICML18}) 
and propagate this bound to the output layer. The robust loss is computed as a cross-entropy loss of verified worse-case outputs ($\ell' = \ell$) 
and then combined with natural prediction loss ($\alpha \neq 0$) as the final loss value during training.

\subsection{Membership Inference Attacks}

For a target machine learning model, the membership inference attacks aim to determine whether a given data point was used to train the model or not 
\cite{shokri_membership_SP17, yeom_privacy_CSF18, salem_membership_NDSS19, nasr_whitebox_privacy_arxiv18, long_privacy_arxiv17, hayes_membership_PETS18}. 
The attack poses a serious privacy risk to the individuals whose data is used for model training, for example in the setting of health analytics.

Shokri et al. \cite{shokri_membership_SP17} design a membership inference attack method based on training an inference model to distinguish 
between predictions on training set members versus non-members.  
To train the inference model, they introduce the \emph{shadow training technique}: 
(1) the adversary first trains multiple ``shadow models'' which simulate the behavior of the target model, 
(2) based on the shadow models' outputs on their own training and test examples, the adversary obtains a labeled (member vs non-member) dataset, and 
(3) finally trains the inference model as a neural network to perform membership inference attack against the target model. 
The input to the inference model is the prediction vector of the target model on a target data record. 

A simpler inference model, such as a linear classifier, can also distinguish significantly vulnerable members from non-members. 
Yeom et al. \cite{yeom_privacy_CSF18} suggest comparing the prediction confidence value of a target example with a threshold 
(learned for example through shadow training). 
Large confidence indicates membership. 
Their results show that such a simple confidence-thresholding method is reasonably effective 
and achieves membership inference accuracy close to that of a complex neural network classifier learned from shadow training.

In this paper, we use this confidence-thresholding membership inference approach in most cases.
Note that when evaluating the privacy leakage with targeted adversarial examples in Section \ref{subsubsec:targeted_adv_design} and Section \ref{subsubsec:targeted_adv}, the confidence-thresholding approach does not apply as there are multiple prediction vectors for each data point. 
Instead, we follow Shokri et al. \cite{shokri_membership_SP17} to train a neural network classifier for membership inference.

\section{Membership Inference Attacks against Robust Models}

In this section, we first present some insights on why training models to be robust against adversarial examples make them more susceptible to membership inference attacks. 
We then formally present our membership inference attacks.

Throughout the paper, we use ``\emph{natural (default) model}'' and ``\emph{robust model}'' to denote the machine learning model 
with natural training algorithm and robust training algorithm, respectively.
We also call the unmodified inputs and adversarially perturbed inputs as ``\emph{benign examples}'' and ``\emph{adversarial examples}''.
When evaluating the model's classification performance, ``\emph{train accuracy}'' and ``\emph{test accuracy}'' are used to 
denote the classification accuracy of benign examples from training and test  sets; 
``\emph{adversarial train accuracy'}' and ``\emph{adversarial test accuracy}'' represent the classification accuracy of adversarial examples from training and test sets; 
``\emph{verified train accuracy}'' and ``\emph{verified test accuracy}'' measure the classification accuracy under the verified worst-case predictions from training and test sets.
Finally, an input example is called ``\emph{secure}'' when it is correctly classified by the model for all adversarial perturbations within the constraint $\mathcal{B}_{\epsilon}$, 
``\emph{insecure}'' otherwise.

The performance of membership inference attacks is highly related to generalization error of target models~\cite{shokri_membership_SP17, yeom_privacy_CSF18}.
An extremely simple attack algorithm can infer membership based on whether or not an input is correctly classified. 
In this case, it is clear that a large gap between the target model's train and test accuracy leads to a significant membership inference attack accuracy 
(as most members are correctly classified, but not the non-members).  
Tsipras et al. \cite{tsipras_tradeoff_ICLR19} and Zhang et al. \cite{zhang_TRADES_ICML19} show that robust training might lead to a drop in test accuracy. 
This is shown based on both empirical and theoretical analysis on toy classification tasks. 
Moreover, the generalization gap can be enlarged for a robust model when evaluating its accuracy on 
adversarial examples~\cite{song_ATDA_ICLR19, schmidt_adv_trained_NIPS18}.
Thus, compared with the natural models, \textbf{the robust models might leak more membership information, due to exhibiting a larger generalization error, 
in both the benign or adversarial settings.}

The performance of membership inference attack is related to the target model's sensitivity with regard to training data \cite{long_privacy_arxiv17}. 
The sensitivity measure is the influence of one data point on the target model's performance by computing its prediction difference, when trained with and without this data point. 
Intuitively, when a training point has a large influence on the target model (high sensitivity),
its model prediction is likely to be different from the model prediction on a test point, and thus the adversary can distinguish its membership more easily.
The robust training algorithms aim to ensure that model predictions remain unchanged for a small area (such as the $l_{\infty}$ ball) around any data point. 
However, in practice, they guarantee this for the training examples, thus, magnifying the influence of the training data on the model.
Therefore, compared with the natural training, \textbf{the robust training algorithms might make the model more susceptible to membership inference attacks, 
by increasing its sensitivity to its training data.}

To validate the above insights, let's take the natural and the robust CIFAR10 classifiers provided by Madry et al. \cite{madry_robust_ICLR18} as an example.
From Figure \ref{fig:loss_for_privacy}, we have seen that compared to the natural model, the robust model has a larger divergence 
between the prediction loss of training data and test data. 
Our fine-grained analysis in Appendix \ref{sec:fine_grained_analysis} further reveals that the large divergence of robust model is highly related to its robustness performance.
Moreover, the robust model incurs a significant generalization error in the adversarial setting, with $96\%$ adversarial train accuracy, and only $47\%$ adversarial test accuracy.
Finally, we will experimentally show in Section \ref{subsubsec:sensitivity} that the robust model is indeed more sensitive with regard to training data.

\subsection{Membership Inference Performance}

\begin{table}[!htb]
\caption{Notations for membership inference attacks against robust machine learning models.
}
\centering
\renewcommand\arraystretch{1.5}
\fontsize{6.7pt}{6.7pt}\selectfont
\begin{tabular}{|c|c|}
\hline
\textbf{Symbol} & \textbf{Description}\\
\hline
\multirow{1}{*}{$F$} & \multirow{1}{*}{Target machine learning model.} \\
\hline
\multirow{1}{*}{$\mathcal{B}_{\epsilon}$} & Adversarial perturbation constraint when training a robust model. \\
\hline
\multirow{1}{*}{$D_{\text{train}}$} & \multirow{1}{*}{Model's training  set.} \\
\hline
\multirow{1}{*}{$D_{\text{test}}$} & \multirow{1}{*}{Model's test  set.} \\
\hline
\multirow{1}{*}{$\bfx$} & \multirow{1}{*}{Benign (unmodified) input example.} \\
\hline
\multirow{1}{*}{$y$} & \multirow{1}{*}{Ground truth label for the input $\bfx$.} \\
\hline
\multirow{1}{*}{$\bfx_{adv}$} & \multirow{1}{*}{Adversarial example generated from $\bfx$.}\\
\hline
\multirow{1}{*}{$\mathcal{V}$} & Robustness verification to compute verified worst-case predictions. \\
\hline
\multirow{1}{*}{$\mathcal{I}$} & Membership inference strategy.\\
\hline
\multirow{1}{*}{$A_{inf}$} & Membership inference accuracy. \\
\hline
\multirow{1}{*}{$ADVT_{inf}$} & Membership inference advantage compared to random guessing.\\
\hline
\end{tabular}
\label{tab:notations}
\end{table}
In this part, we describe the membership inference attack and its performance formally, with notations listed in Table \ref{tab:notations}.
For a neural network model $F$ (we skip its parameter $\theta$ for simplicity) that is robustly trained with the adversarial constraint $\mathcal{B}_{\epsilon}$, 
the membership inference attack aims to determine whether a given input example $\bfz = (\bfx, y)$ is in its training set $D_{\text{train}}$ or not.  
We denote the inference strategy adopted by the adversary as $\mathcal{I}(F, \mathcal{B}_{\epsilon}, \bfz)$, which codes members as 1, and non-members as 0.

We use the fraction of correct membership predictions, as the metric to evaluate membership inference accuracy.
We use a test set $D_{\text{test}}$ which does not overlap with the training set, to represent non-members. 
We sample a random data point ($\bfx$, $y$) from either $D_{\text{train}}$ or $D_{\text{test}}$ with an equal $50\%$ probability, 
to test the membership inference attack. We measure the membership inference accuracy as follows.
\begin{equation}\label{eq:inf_accuracy}
\begin{aligned}
{A}_{inf}( & F, \mathcal{B}_{\epsilon}, \mathcal{I}) =  \frac{\sum_{\bfz \in D_{\text{train}}} \mathcal{I}(F, \mathcal{B}_{\epsilon}, \bfz)} {2 \cdot |D_{\text{train}}|}  + \frac{\sum_{\bfz \in D_{\text{test}}} 1 - \mathcal{I}(F, \mathcal{B}_{\epsilon}, \bfz)} {2 \cdot |D_{\text{test}}|}, 
\end{aligned}
\end{equation}
where $|\cdot|$ measures the size of a dataset.

The membership inference accuracy evaluates the probability that the adversary can guess correctly whether an input is from training set or test set. Note that a random guessing strategy will lead to a $50\%$ inference accuracy.
To further measure the effectiveness of our membership inference strategy, we also use the notion of membership inference advantage proposed by Yeom et al. \cite{yeom_privacy_CSF18}, which is defined as the increase in inference accuracy over random guessing (multiplied by $2$).
\begin{equation}\label{eq:inf_advantage}
{ADVT}_{inf} = 2 \times (A_{inf} - 0.5)
\end{equation}

\subsection{Exploiting the Model's Predictions on Benign Examples}

We adopt a confidence-thresholding inference strategy due to its simplicity and effectiveness \cite{yeom_privacy_CSF18}: 
an input $(\bfx, y)$ is inferred as member if its prediction confidence $F(\bfx)_{y}$ is larger than a preset threshold value.
We denote this inference strategy as $\mathcal{I}_{\boldsymbol{\mathrm{B}}}$ since it relies on the benign examples' predictions. 
We have the following expressions for this inference strategy and its inference accuracy.
\begin{equation}\label{eq:nat_conf_based_inf}
\begin{aligned}
\mathcal{I}_{\boldsymbol{\mathrm{B}}}(F, \mathcal{B}_{\epsilon}, (\bfx, y)) = & \, \mathbbm{1}\{F(\bfx)_{y} \geq \tau_{\boldsymbol{\mathrm{B}}}\}\\
{A}_{inf}(F, \mathcal{B}_{\epsilon}, \mathcal{I}_{\boldsymbol{\mathrm{B}}}) = & \, \frac{1}{2} + \frac{1}{2} \cdot (\frac{\sum_{\bfz \in D_{\text{train}}} \mathbbm{1}\{F(\bfx)_{y} \geq \tau_{\boldsymbol{\mathrm{B}}}\}}{|D_{\text{train}}|} \\
& \, - \frac{\sum_{\bfz \in D_{\text{test}}} \mathbbm{1}\{F(\bfx)_{y} \geq \tau_{\boldsymbol{\mathrm{B}}}\}}{|D_{\text{test}}|}),
\end{aligned}
\end{equation}
where $\mathbbm{1}\{\cdot\}$ is the indicator function and the last two terms are the values of complementary cumulative distribution functions 
of training examples' and test examples' prediction confidences, at the point of threshold $\tau_{\boldsymbol{\mathrm{B}}}$, respectively.
In our experiments, we evaluate the worst case inference risks by choosing $\tau_{\boldsymbol{\mathrm{B}}}$ to achieve the highest inference accuracy, 
i.e., maximizing the gap between two complementary cumulative distribution function values.
In practice, an adversary can learn the threshold via the shadow training technique \cite{shokri_membership_SP17}.

This inference strategy $\mathcal{I}_{\boldsymbol{\mathrm{B}}}$ does not leverage the adversarial constraint $\mathcal{B}_{\epsilon}$ of the robust model.
Intuitively, the robust training algorithm learns to make smooth predictions around training examples.
In this paper, we observe that such smooth predictions around training examples may not generalize well to test examples and we can leverage this property to perform stronger membership inference attacks.
Based on this observation, we propose two new membership inference strategies against robust models by taking $\mathcal{B}_{\epsilon}$ into consideration.

\subsection{Exploiting the Model's Predictions on Adversarial Examples}

Our first new inference strategy is to generate an (untargeted) adversarial example $\bfx_{adv}$ for input $(\bfx, y)$ under the constraint $\mathcal{B}_{\epsilon}$, 
and use a threshold on the model's prediction confidence on $\bfx_{adv}$.  
We have following expression for this strategy $\mathcal{I}_{\boldsymbol{\mathrm{A}}}$ and its inference accuracy.
\begin{equation}\label{eq:adv_conf_based_inf}
\begin{aligned}
\mathcal{I}_{\boldsymbol{\mathrm{A}}}(F, \mathcal{B}_{\epsilon}, (\bfx, y)) = & \, \mathbbm{1}\{F(\bfx_{adv})_{y} \geq \tau_{\boldsymbol{\mathrm{A}}}\}\\
{A}_{inf}(F, \mathcal{B}_{\epsilon}, \mathcal{I}_{\boldsymbol{\mathrm{A}}}) = & \, \frac{1}{2} + \frac{1}{2} \cdot (\frac{\sum_{\bfz \in D_{\text{train}}} \mathbbm{1}\{F(\bfx_{adv})_{y} \geq \tau_{\boldsymbol{\mathrm{A}}}\}}{|D_{\text{train}}|} \\
& \, - \frac{\sum_{\bfz \in D_{\text{test}}} \mathbbm{1}\{F(\bfx_{adv})_{y} \geq \tau_{\boldsymbol{\mathrm{A}}}\}}{|D_{\text{test}}|})
\end{aligned}
\end{equation}

We use the PGD attack method shown in Equation \eqref{eq:PGD} to obtain $\bfx_{adv}$.
Similarly, we choose the preset threshold $\tau_{\boldsymbol{\mathrm{A}}}$ to achieve the highest inference accuracy, 
i.e., maximizing the gap between two complementary cumulative distribution functions of prediction confidence on adversarial train and test examples.

To perform membership inference attacks with the strategy $\mathcal{I}_{\boldsymbol{\mathrm{A}}}$, we need to specify the perturbation constraint $\mathcal{B}_{\epsilon}$.
For our experimental evaluations in Section \ref{sec:results_empirical_robust} and Section \ref{sec:results_verifiably_robust}, we use the same perturbation constraint $\mathcal{B}_{\epsilon}$ as in the robust training process, which is assumed to be prior knowledge of the adversary.
We argue that this assumption is reasonable following Kerckhoffs's principle \cite{kerckhoffs1883cryptographic, shannon1949communication}.
In Section \ref{subsec:different-lp-balls}, we measure privacy leakage when the robust model's perturbation constraint is unknown.

\subsubsection{Targeted adversarial examples}
\label{subsubsec:targeted_adv_design}

We extend the attack to exploiting targeted adversarial examples.
Targeted adversarial examples contain information about distance of the benign input to each label's decision boundary, 
and are expected to leak more membership information than the untargeted adversarial example 
which only contains information about distance to a nearby label's decision boundary.

We adapt the PGD attack method to find targeted adversarial examples (Equation \eqref{eq:tar-adv-example}) by iteratively minizing the targeted cross-entropy loss.
\begin{equation}\label{eq:PGD-target}
\tilde{\bfx}^{t+1} = \Pi_{\mathcal{B}_{\epsilon}(\bfx)} [\tilde{\bfx}^{t} - \eta \cdot \text{sign}(\, \nabla_{\tilde{\bfx}^{t}} \ell(F_{\theta},(\tilde{\bfx}^{t},y') ) \,) ]
\end{equation}

The confidence thresholding inference strategy does not apply for targeted adversarial examples because there exist $k-1$ targeted adversarial examples 
(we have $k-1$ incorrect labels) for each input.
Instead, following Shokri et al. \cite{shokri_membership_SP17}, we train a binary inference classifier for each class label to perform the membership inference attack. 
For each class label, we  first choose a fraction of training and test points and generate corresponding targeted adversarial examples. 
Next, we compute model predictions on the targeted adversarial examples, and use them to train the membership inference classifier. 
Finally, we perform inference attacks using the remaining training and test points.

\subsection{Exploiting the Verified Worst-Case Predictions on Adversarial Examples}

Our attacks above generate adversarial examples using the \emph{heuristic} strategy of projected gradient descent. 
Next, we leverage \emph{verification} techniques $\mathcal{V}$ used by the verifiably defended models 
\cite{wong_robust_ICML18, mirman_provable_ICML18, gowal_IBP_secml18} to obtain the input's worst-case predictions 
under the adversarial constraint $\mathcal{B}_{\epsilon}$. 
We use the input's worst-case prediction confidence to  predict its membership. 
The expressions for this strategy $\mathcal{I}_{\boldsymbol{\mathrm{V}}}$ and its inference accuracy are as follows.
\begin{equation}\label{eq:ver_conf_based_inf}
\begin{aligned}
\mathcal{I}_{\boldsymbol{\mathrm{V}}}(F, \mathcal{B}_{\epsilon}, (\bfx, y)) = & \, \mathbbm{1}\{\mathcal{V}(F(\tilde{\bfx})_{y}, \mathcal{B}_{\epsilon}) \geq \tau_{\boldsymbol{\mathrm{V}}}\}\\
{A}_{inf}(F, \mathcal{B}_{\epsilon}, \mathcal{I}_{\boldsymbol{\mathrm{V}}}) = & \, \frac{1}{2} + \frac{1}{2} \cdot (\frac{\sum_{\bfz \in D_{\text{train}}} \mathcal{V}(F(\tilde{\bfx})_{y}, \mathcal{B}_{\epsilon}) \geq \tau_{\boldsymbol{\mathrm{V}}}\}}{|D_{\text{train}}|} \\
& \, - \frac{\sum_{\bfz \in D_{\text{test}}} \mathcal{V}(F(\tilde{\bfx})_{y}, \mathcal{B}_{\epsilon}) \geq \tau_{\boldsymbol{\mathrm{V}}}\}}{|D_{\text{test}}|}),
\end{aligned}
\end{equation}
where $\mathcal{V}(F(\tilde{\bfx})_{y}, \mathcal{B}_{\epsilon})$ returns the verified worst-case prediction confidence for all examples $\tilde{\bfx}$ 
satisfying the adversarial perturbation constraint $\tilde{\bfx} \in \mathcal{B}_{\epsilon}(\bfx)$, 
and $\tau_{\boldsymbol{\mathrm{V}}}$ is chosen in a similar manner as our previous two inference strategies.

Note that different verifiable defenses adopt different verification methods $\mathcal{V}$. 
Our inference strategy $\mathcal{I}_{\boldsymbol{\mathrm{V}}}$ needs to use the same verification method which is used in the target model's verifiably robust training process.
Again, we argue that it is reasonable to assume that an adversary has knowledge about the verification method $\mathcal{V}$ and the perturbation constraint $\mathcal{B}_{\epsilon}$, following Kerckhoffs's principle \cite{kerckhoffs1883cryptographic, shannon1949communication}.

\section{Experiment Setup}\label{sec:setup}

In this section, we describe the datasets, neural network architectures, and corresponding adversarial perturbation constraints that we use in our experiments. 
Throughout the paper, we focus on the $l_{\infty}$ perturbation constraint: $\mathcal{B}_{\epsilon}(\bfx) = \{\bfx' \, | \, \lVert \bfx'- \bfx \rVert_{\infty} \leq \epsilon \}$.
The detailed architectures are summarized in Appendix \ref{append:NN_Models}. 
Our code is publicly available at \url{https://github.com/inspire-group/privacy-vs-robustness}.

\noindent \textbf{Yale Face.}
The extended Yale Face database B is used to train face recognition models, and contains gray scale face images of $38$ subjects under various lighting conditions 
\cite{georghiades_yaleface_01, lee_yaleface_05}.
We use the cropped version of this dataset, where all face images are aligned and cropped to have the dimension of $168 \times 192$. 
In this version, each subject has $64$ images with the same frontal poses under different lighting conditions, among which $18$ images were corrupted during the image acquisition, 
leading to 2,414 images in total \cite{lee_yaleface_05}.
In our experiments, we select $50$ images for each subject to form the training  set (total size is 1,900 images), and use the remaining 514 images as the test set.

For the model architecture, we use a convolutional neural network (CNN) with the convolution kernel size $3 \times 3$, 
as suggested by Simonyan et al. \cite{simonyan_DNN_arxiv14}.
The CNN model contains 4 blocks with different numbers of output channels $(8, 16, 32, 64)$, and each block contains two convolution layers. 
The first layer uses a stride of $1$ for convolutions, and the second layer uses a stride of $2$. 
There are two fully connected layers after the convolutional layers, each containing $200$ and $38$ neurons.
When training the robust models, we set the $l_{\infty}$ perturbation budget ($\epsilon$) to be $8/255$.

\noindent \textbf{Fashion-MNIST.}
This dataset consists of a training set of 60,000 examples and a test set of 10,000 examples \cite{xiao_fmnist_17}. 
Each example is a $28 \times 28$ grayscale image, associated with a class label from 10 fashion products, such as shirt, coat, sneaker.

Similar to Yale Face, we also adopt a CNN architecture with the convolution kernel size $3 \times 3$. 
The model contains 2 blocks with output channel numbers $(256, 512)$, and each block contains three convolution layers. 
The first two layers both use a stride of $1$, while the last layer uses a stride of $2$. 
Two fully connected layers are added at the end, with $200$ and $10$ neurons, respectively.
When training the robust models, we set the $l_{\infty}$ perturbation budget ($\epsilon$) to be $0.1$.

\noindent \textbf{CIFAR10.}
This dataset is composed of $32 \times 32$ color images in 10 classes, with 6,000 images per class. In total, there are 50,000 training images and 10,000 test images.

We use the wide ResNet architecture \cite{Zagoruyko_WRN_BMVC16} to train a CIFAR10 classifier, following Madry et al. \cite{madry_robust_ICLR18}. 
It contains 3 groups of residual layers with output channel numbers (160, 320, 640) and 5 residual units for each group. 
One fully connected layer with $10$ neurons is added at the end.
When training the robust models, we set the $l_{\infty}$ perturbation budget ($\epsilon$) to be $8/255$.

\section{Membership Inference Attacks against Empirically Robust Models}\label{sec:results_empirical_robust}
In this section we discuss membership inference attacks against 3 empirical defense methods: 
PGD-based adversarial training (PGD-Based Adv-Train) \cite{madry_robust_ICLR18}, distributional adversarial training (Dist-Based Adv-Train) \cite{sinha_PAT_ICLR18}, 
and difference-based adversarial training (Diff-Based Adv-Train) \cite{zhang_TRADES_ICML19}.
We train the robust models against the $l_{\infty}$ adversarial constraint on the Yale Face dataset, the Fashion-MNIST dataset, and the CIFAR10 dataset, 
with neural network architecture as described in Section \ref{sec:setup}. 
Following previous work \cite{athalye_adv_ICML18, madry_robust_ICLR18, wong_robust_ICML18, mirman_provable_ICML18}, 
the perturbation budget $\epsilon$ values are set to be $8/255$, $0.1$, and $8/255$ on three datasets, respectively.
For the empirically robust model, as explained in Section \ref{subsec:adv_attack}, there is no verification process to obtain robustness guarantee. 
Thus the membership inference strategy $\mathcal{I}_{\boldsymbol{\mathrm{V}}}$ does not apply here.

We first present an overall analysis that compares membership inference accuracy for natural models and robust models 
using multiple inference strategies across multiple datasets. 
We then present a deeper analysis of membership inference attacks against the PGD-based adversarial training defense.

\subsection{Overall Results}

\begin{table}[!htb]
\caption{Membership inference attacks against natural and empirically robust models \cite{madry_robust_ICLR18, sinha_PAT_ICLR18, zhang_TRADES_ICML19} 
on the Yale Face dataset with a $l_{\infty}$ perturbation constraint $\epsilon=8/255$. 
Based on Equation \eqref{eq:inf_advantage}, the natural model has an inference advantage of $11.70\%$, while the robust model has an inference advantage up to $37.66\%$.
}
\centering
\renewcommand\arraystretch{1.3}
\fontsize{6.7pt}{6.7pt}\selectfont
\begin{tabular}{|c|c|c|c|c|c|c|}
\hline
Training & train & test & adv-train & adv-test & inference & inference \\
method & acc & acc & acc & acc & acc ($\mathcal{I}_{\boldsymbol{\mathrm{B}}}$) & acc ($\mathcal{I}_{\boldsymbol{\mathrm{A}}}$) \\
\hline
\multirow{2}{*}{Natural} & \multirow{2}{*}{100\%} & \multirow{2}{*}{98.25\%}  & \multirow{2}{*}{4.53\%} & \multirow{2}{*}{2.92\%} & \multirow{2}{*}{\textbf{55.85\%}} & \multirow{2}{*}{54.27\%}  \\
 & & & & & & \\
\hline
PGD-Based & \multirow{2}{*}{99.89\%} & \multirow{2}{*}{96.69\%}  & \multirow{2}{*}{99.00\%} & \multirow{2}{*}{77.63\%} & \multirow{2}{*}{61.69\%} & \multirow{2}{*}{\textbf{68.83\%}} \\
Adv-Train \cite{madry_robust_ICLR18} & & & & & &  \\
\hline
Dist-Based  & \multirow{2}{*}{99.58\%} & \multirow{2}{*}{93.77\%} & \multirow{2}{*}{83.26\%} & \multirow{2}{*}{55.06\%} & \multirow{2}{*}{62.23\%} & \multirow{2}{*}{\textbf{64.07\%}}\\
Adv-Train \cite{sinha_PAT_ICLR18} & & & & & &  \\
\hline
Diff-Based & \multirow{2}{*}{99.53\%} & \multirow{2}{*}{93.77\%}  & \multirow{2}{*}{99.42\%} & \multirow{2}{*}{83.85\%} & \multirow{2}{*}{58.06\%} & \multirow{2}{*}{\textbf{65.59\%}}  \\
Adv-Train \cite{zhang_TRADES_ICML19} & & & & & &  \\
\hline
\end{tabular}
\label{tab:empirical_defenses_privacy_YALEFACE}
\end{table}

\begin{table}[!htb]
\caption{Membership inference attacks against natural and empirically robust models \cite{madry_robust_ICLR18, sinha_PAT_ICLR18, zhang_TRADES_ICML19} 
on the Fashion-MNIST dataset with a $l_{\infty}$ perturbation constraint $\epsilon=0.1$. 
Based on Equation \eqref{eq:inf_advantage}, the natural model has an inference advantage of $14.24\%$, while the robust model has an inference advantage up to $28.98\%$.
}
\centering
\renewcommand\arraystretch{1.3}
\fontsize{6.7pt}{6.7pt}\selectfont
\begin{tabular}{|c|c|c|c|c|c|c|}
\hline
Training & train & test & adv-train & adv-test & inference & inference \\
method & acc & acc & acc & acc & acc ($\mathcal{I}_{\boldsymbol{\mathrm{B}}}$) & acc ($\mathcal{I}_{\boldsymbol{\mathrm{A}}}$) \\
\hline
\multirow{2}{*}{Natural} & \multirow{2}{*}{100\%} & \multirow{2}{*}{92.18\%}  & \multirow{2}{*}{4.35\%} & \multirow{2}{*}{4.14\%} & \multirow{2}{*}{\textbf{57.12\%}} & \multirow{2}{*}{50.95\%}  \\
 & & & & & & \\
\hline
PGD-Based & \multirow{2}{*}{99.93\%} & \multirow{2}{*}{90.88\%}  & \multirow{2}{*}{96.91\%} & \multirow{2}{*}{68.06\%} & \multirow{2}{*}{58.32\%} & \multirow{2}{*}{\textbf{64.49\%}} \\
Adv-Train \cite{madry_robust_ICLR18} & & & & & &  \\
\hline
Dist-Based  & \multirow{2}{*}{97.98\%} & \multirow{2}{*}{90.62\%} & \multirow{2}{*}{67.63\%} & \multirow{2}{*}{51.61\%} & \multirow{2}{*}{57.35\%} & \multirow{2}{*}{\textbf{59.49\%}}\\
Adv-Train \cite{sinha_PAT_ICLR18} & & & & & &  \\
\hline
Diff-Based & \multirow{2}{*}{99.35\%} & \multirow{2}{*}{90.92\%}  & \multirow{2}{*}{90.13\%} & \multirow{2}{*}{72.40\%} & \multirow{2}{*}{57.02\%} & \multirow{2}{*}{\textbf{58.83\%}}  \\
Adv-Train \cite{zhang_TRADES_ICML19} & & & & & &  \\
\hline
\end{tabular}
\label{tab:empirical_defenses_privacy_FMNIST}
\end{table}

\begin{table}[!htb]
\caption{Membership inference attacks against natural and empirically robust models \cite{madry_robust_ICLR18, sinha_PAT_ICLR18, zhang_TRADES_ICML19} 
on the CIFAR10 dataset with a $l_{\infty}$ perturbation constraint $\epsilon=8/255$. 
Based on Equation \eqref{eq:inf_advantage}, the natural model has an inference advantage of $14.86\%$, while the robust model has an inference advantage up to $51.34\%$.
}
\centering
\renewcommand\arraystretch{1.3}
\fontsize{6.7pt}{6.7pt}\selectfont
\begin{tabular}{|c|c|c|c|c|c|c|}
\hline
Training & train & test & adv-train & adv-test & inference & inference \\
method & acc & acc & acc & acc & acc ($\mathcal{I}_{\boldsymbol{\mathrm{B}}}$) & acc ($\mathcal{I}_{\boldsymbol{\mathrm{A}}}$) \\
\hline
\multirow{2}{*}{Natural} & \multirow{2}{*}{100\%} & \multirow{2}{*}{95.01\%} & \multirow{2}{*}{0\%} & \multirow{2}{*}{0\%} & \multirow{2}{*}{\textbf{57.43\%}} & \multirow{2}{*}{50.86\%}   \\
 & & & & & & \\
\hline
PGD-Based & \multirow{2}{*}{99.99\%} & \multirow{2}{*}{87.25\%} & \multirow{2}{*}{96.08\%} & \multirow{2}{*}{46.61\%} & \multirow{2}{*}{74.89\%} &  \multirow{2}{*}{\textbf{75.67\%}} \\
Adv-Train \cite{madry_robust_ICLR18} & & & & & &  \\
\hline
Dist-Based  & \multirow{2}{*}{100\%} & \multirow{2}{*}{90.10\%} & \multirow{2}{*}{40.56\%} & \multirow{2}{*}{25.92\%} & \multirow{2}{*}{\textbf{67.16\%}} & \multirow{2}{*}{64.24\%}\\
Adv-Train \cite{sinha_PAT_ICLR18} & & & & & &  \\
\hline
Diff-Based & \multirow{2}{*}{99.50\%} & \multirow{2}{*}{87.99\%}  & \multirow{2}{*}{76.06\%} & \multirow{2}{*}{46.50\%} & \multirow{2}{*}{61.18\%} & \multirow{2}{*}{\textbf{67.08\%}}  \\
Adv-Train \cite{zhang_TRADES_ICML19} & & & & & &  \\
\hline
\end{tabular}
\label{tab:empirical_defenses_privacy_CIFAR10}
\end{table}

The membership inference attack results against natural models and empirically robust models \cite{madry_robust_ICLR18, sinha_PAT_ICLR18, zhang_TRADES_ICML19} 
are presented in Table \ref{tab:empirical_defenses_privacy_YALEFACE}, Table  \ref{tab:empirical_defenses_privacy_FMNIST} and 
Table  \ref{tab:empirical_defenses_privacy_CIFAR10}, 
where ``acc'' stands for accuracy, while ``adv-train acc'' and ``adv-test acc'' report adversarial accuracy under PGD attacks as shown in Equation \eqref{eq:PGD}.

According to these results, {\bf all three empirical defense methods will make the model more susceptible to membership inference attacks}: 
compared with natural models, robust models increase the membership inference advantage by up to $3.2\times$, $2\times$, and $3.5 \times$, 
for Yale Face, Fashion-MNIST, and CIFAR10, respectively.

We also find that {\bf for robust models, membership inference attacks based on adversarial example's prediction confidence ($\mathcal{I}_{\boldsymbol{\mathrm{A}}}$) 
have higher inference accuracy than the inference attacks based on benign example's prediction confidence ($\mathcal{I}_{\boldsymbol{\mathrm{B}}}$) in most cases. 
On the other hand, for natural models, inference attacks based on benign examples' prediction confidence lead to higher inference accuracy values.}
This happens because our inference strategies rely on the difference between confidence distribution of training points and that of test points.
For robust models, most of training points are (empirically) secure against adversarial examples, 
and adversarial perturbations do not significantly decrease the confidence on them. 
However, the test set contains more insecure points, and thus adversarial perturbations will enlarge the gap 
between confidence distributions of training examples and test examples, leading to a higher inference accuracy. 
For natural models, the use of adversarial examples will decrease the confidence distribution gap, 
since almost all training points and test points are not secure with adversarial perturbations.
The only exception is Dist-Based Adv-Train CIFAR10 classifier, where inference accuracy with strategy $\mathcal{I}_{\boldsymbol{\mathrm{B}}}$ is higher, 
which can be explained by the poor robustness performance of the model: around $60\%$ training examples are insecure. 
Thus, adversarial perturbations will decrease the confidence distribution gap between training examples and test examples in this specific scenario.

\subsection{Detailed Membership Inference Analysis of PGD-Based Adversarial Training} \label{subsec:PGD-adv-train}

In this part, we perform a detailed analysis of membership inference attacks against PGD-based adversarial training defense method \cite{madry_robust_ICLR18} 
by using the CIFAR10 classifier as an example.
We first perform a sensitivity analysis on both natural and robust models to show that the robust model is more sensitive with regard to training data compared to the natural model.
We then investigate the relation between privacy leakage and model properties, including robustness generalization, adversarial perturbation constraint and model capacity.
We finally show that the predictions of targeted adversarial examples can further enhance the membership inference advantage.

\subsubsection{Sensitivity Analysis} \label{subsubsec:sensitivity}

\begin{figure}[!ht]
	\centering
	\includegraphics[width=\linewidth]{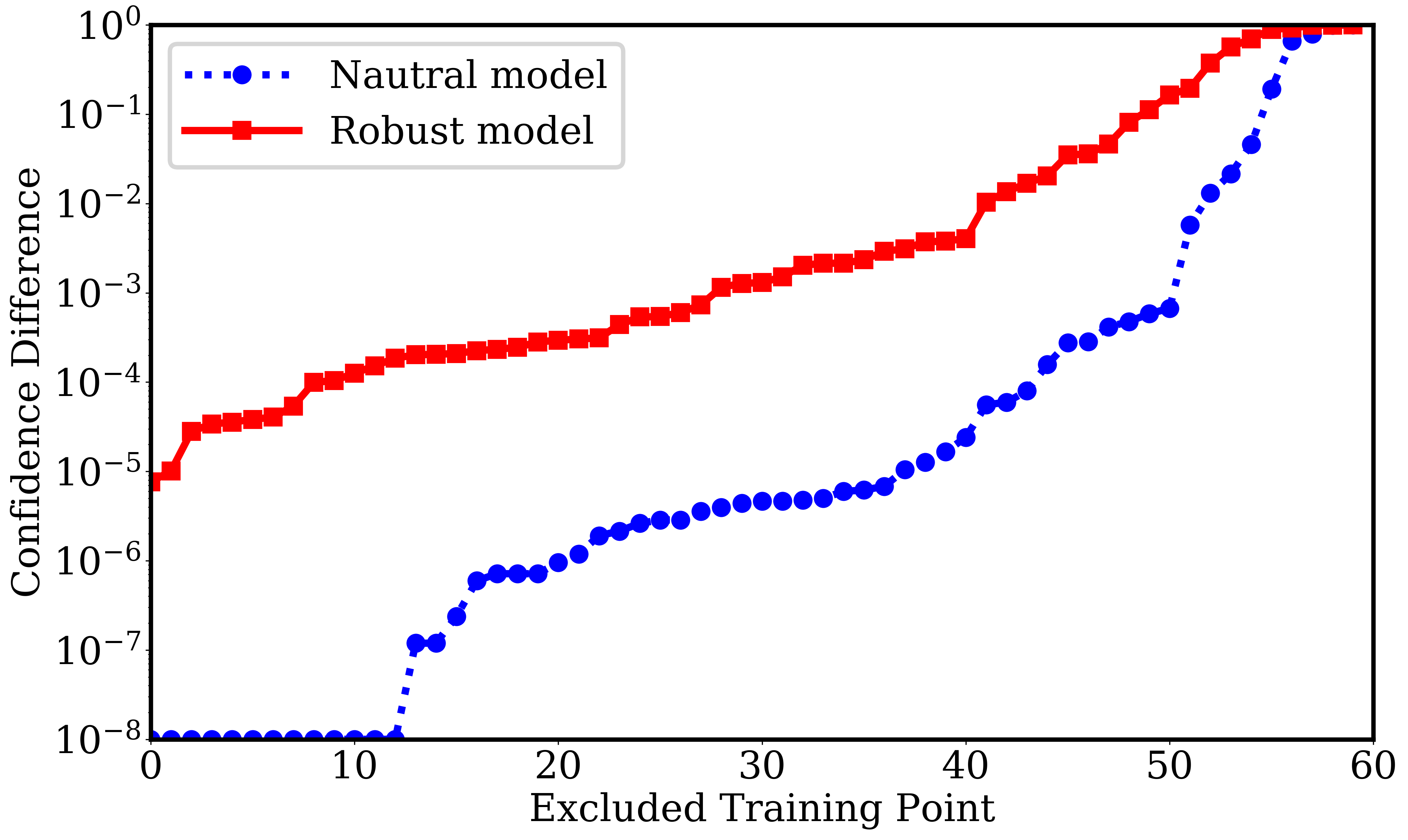}
	\caption{Sensitivity analysis of both robust~\cite{madry_robust_ICLR18} and natural CIFAR10 classifiers.    x-axis denotes the excluded training point id number (sorted by sensitivity) during the retraining process, and y-axis denotes the difference in prediction confidence between the original model and the retrained model (measuring model sensitivity). 
	 The robust model is more sensitive to the training data compared to the natural model.
	}
	\label{fig:sensitivity}
\end{figure}

In the sensitivity analysis, we remove sample CIFAR10 training points from the training set, perform retraining of the models, 
and compute the performance difference between the original model and retrained model.

We excluded 10 training points (one for each class label) and retrained the model.  
We computed the sensitivity of each excluded point as the difference between its prediction confidence in the retrained model and the original model.
We obtained the sensitivity metric for 60 training points by retraining the classifier 6 times.
Figure \ref{fig:sensitivity} depicts the sensitivity values for the 60 training points (in ascending order) for both robust and natural models. 
We can see that compared to the natural model, \textbf{the robust model is indeed more sensitive to the training data, thus leaking more membership information.}

\subsubsection{Privacy risk with robustness generalization}\label{subsubsec:privacy_generalization}

We perform the following experiment to demonstrate the relation between privacy risk and robustness generalization. 
Recall that in the approach of Madry et al.~\cite{madry_robust_ICLR18}, adversarial examples are generated from \emph{all} training points during the robust training process. 
In our experiment, we modify the above defense approach to 
(1) leverage adversarial examples from a \emph{subset} of the CIFAR10 training data to compute the robust prediction loss, and 
(2) leverage the remaining subset of training points as benign inputs to compute the natural prediction loss.

\begin{table}[!htb]
\caption{Mixed PGD-based adversarial training experiments \cite{madry_robust_ICLR18} on CIFAR10 dataset with a $l_{\infty}$ perturbation constraint $\epsilon=8/255$. 
During the training process, part of the training set, whose ratio is denoted by adv-train ratio, is used to compute robust loss, 
and the remaining part of the training set is used to compute natural loss.
}
\centering
\renewcommand\arraystretch{1.3}
\fontsize{6.7pt}{6.7pt}\selectfont
\begin{tabular}{|c|c|c|c|c|c|c|}
\hline
Adv-train & train & test & adv-train & adv-test & inference & inference \\
ratio & acc & acc & acc & acc & acc ($\mathcal{I}_{\boldsymbol{\mathrm{B}}}$) & acc ($\mathcal{I}_{\boldsymbol{\mathrm{A}}}$) \\
\hline
\multirow{2}{*}{0} & \multirow{2}{*}{100\%} & \multirow{2}{*}{95.01\%} & \multirow{2}{*}{0\%} & \multirow{2}{*}{0\%} & \multirow{2}{*}{\textbf{57.43\%}} & \multirow{2}{*}{50.85\%}   \\
 & & & & & & \\
\hline
\multirow{2}{*}{1/2} & \multirow{2}{*}{100\%} & \multirow{2}{*}{87.78\%} & \multirow{2}{*}{75.85\%} & \multirow{2}{*}{43.23\%} & \multirow{2}{*}{\textbf{67.20\%}} & \multirow{2}{*}{66.36\%}\\
 & & & & & &  \\
\hline
\multirow{2}{*}{3/4} & \multirow{2}{*}{100\%} & \multirow{2}{*}{86.68\%}  & \multirow{2}{*}{88.34\%} & \multirow{2}{*}{45.66\%} & \multirow{2}{*}{71.07\%} & \multirow{2}{*}{\textbf{72.22\%}}  \\
& & & & & &  \\
\hline
\multirow{2}{*}{1} & \multirow{2}{*}{99.99\%} & \multirow{2}{*}{87.25\%}  & \multirow{2}{*}{96.08\%} & \multirow{2}{*}{46.61\%} & \multirow{2}{*}{74.89\%} & \multirow{2}{*}{\textbf{75.67\%}}  \\
& & & & & &  \\
\hline
\end{tabular}
\label{tab:mixed_adv_train}
\end{table}

The membership inference attack results are summarized in Table \ref{tab:mixed_adv_train}, 
where the first column lists the ratio of training points used for computing robust loss.
We can see that {\bf as more training points are used for computing the robust loss, the membership inference accuracy increases}, 
due to the larger gap between adv-train accuracy 
and adv-test accuracy. 

\subsubsection{Privacy risk with model perturbation budget}
Next, we explore the relationship between membership inference and the adversarial perturbation budget $\epsilon$, 
which controls the maximum absolute value of adversarial perturbations during robust training process.

\begin{table}[!htb]
\caption{Membership inference attacks against robust CIFAR10 classifiers \cite{madry_robust_ICLR18} with varying adversarial perturbation budgets.
}
\centering
\renewcommand\arraystretch{1.3}
\fontsize{6.7pt}{6.7pt}\selectfont
\begin{tabular}{|c|c|c|c|c|c|c|}
\hline
Perturbation & train & test & adv-train & adv-test & inference & inference \\
budget ($\epsilon$) & acc & acc & acc & acc & acc ($\mathcal{I}_{\boldsymbol{\mathrm{B}}}$) & acc ($\mathcal{I}_{\boldsymbol{\mathrm{A}}}$) \\
\hline
\multirow{2}{*}{2/255} & \multirow{2}{*}{100\%} & \multirow{2}{*}{93.74\%} & \multirow{2}{*}{99.99\%} & \multirow{2}{*}{82.20\%} & \multirow{2}{*}{64.48\%} & \multirow{2}{*}{\textbf{66.54\%}}\\
 & & & & & &  \\
\hline
\multirow{2}{*}{4/255} & \multirow{2}{*}{100\%} & \multirow{2}{*}{91.19\%}  & \multirow{2}{*}{99.89\%} & \multirow{2}{*}{70.03\%} & \multirow{2}{*}{69.44\%} & \multirow{2}{*}{\textbf{72.43\%}}  \\
& & & & & &  \\
\hline
\multirow{2}{*}{8/255} & \multirow{2}{*}{99.99\%} & \multirow{2}{*}{87.25\%}  & \multirow{2}{*}{96.08\%} & \multirow{2}{*}{46.61\%} & \multirow{2}{*}{74.89\%} & \multirow{2}{*}{\textbf{75.67\%}}  \\
& & & & & &  \\
\hline
\end{tabular}
\label{tab:privacy_over_epsilon}
\end{table}

We performed the robust training \cite{madry_robust_ICLR18} for three CIFAR10 classifiers with varying adversarial perturbation budgets, 
and show the result in Table~\ref{tab:privacy_over_epsilon}. 
Note that a model trained with a larger $\epsilon$ is more robust since it can defend against larger adversarial perturbations.
From Table \ref{tab:privacy_over_epsilon}, we can see that \textbf{more robust models leak more information about the training data}.
With a larger $\epsilon$ value, the robust model relies on a larger $l_{\infty}$ ball around each training point, leading to a higher membership inference attack accuracy.

\subsubsection{Privacy risk with model capacity} \label{subsubsec:PGD_defense_capacity}

Madry et al. \cite{madry_robust_ICLR18} have observed that compared with natural training, robust training requires a significantly larger model capacity 
(e.g., deeper neural network architectures and more convolution filters) to obtain high robustness.
In fact, we can think of the robust training approach as adding more ``virtual training points'', 
which are within the $l_{\infty}$ ball around original training points. Thus the model capacity needs to be large enough to fit well on the larger ``virtual training  set''.

\begin{figure}[!ht]
	\centering
	\begin{subfigure}[t]{0.95\linewidth}
		\raggedleft
		\includegraphics[width=\linewidth]{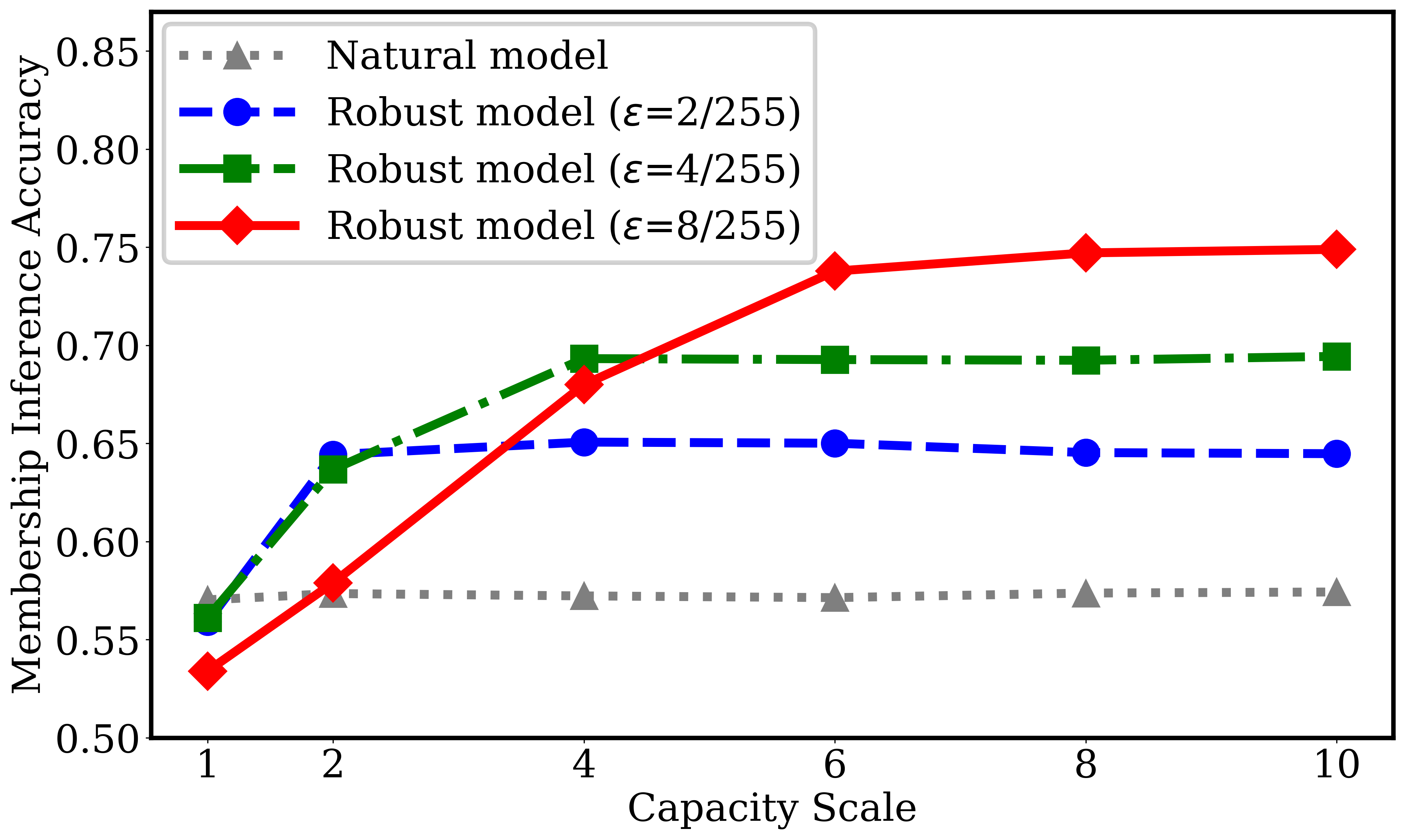}
		\caption{Membership inference attacks against models with different model capacities.}
		\label{fig:infer_capacity}
	\end{subfigure}\hfill
	\begin{subfigure}[t]{0.95\linewidth}
		\raggedright
		\includegraphics[width=\linewidth]{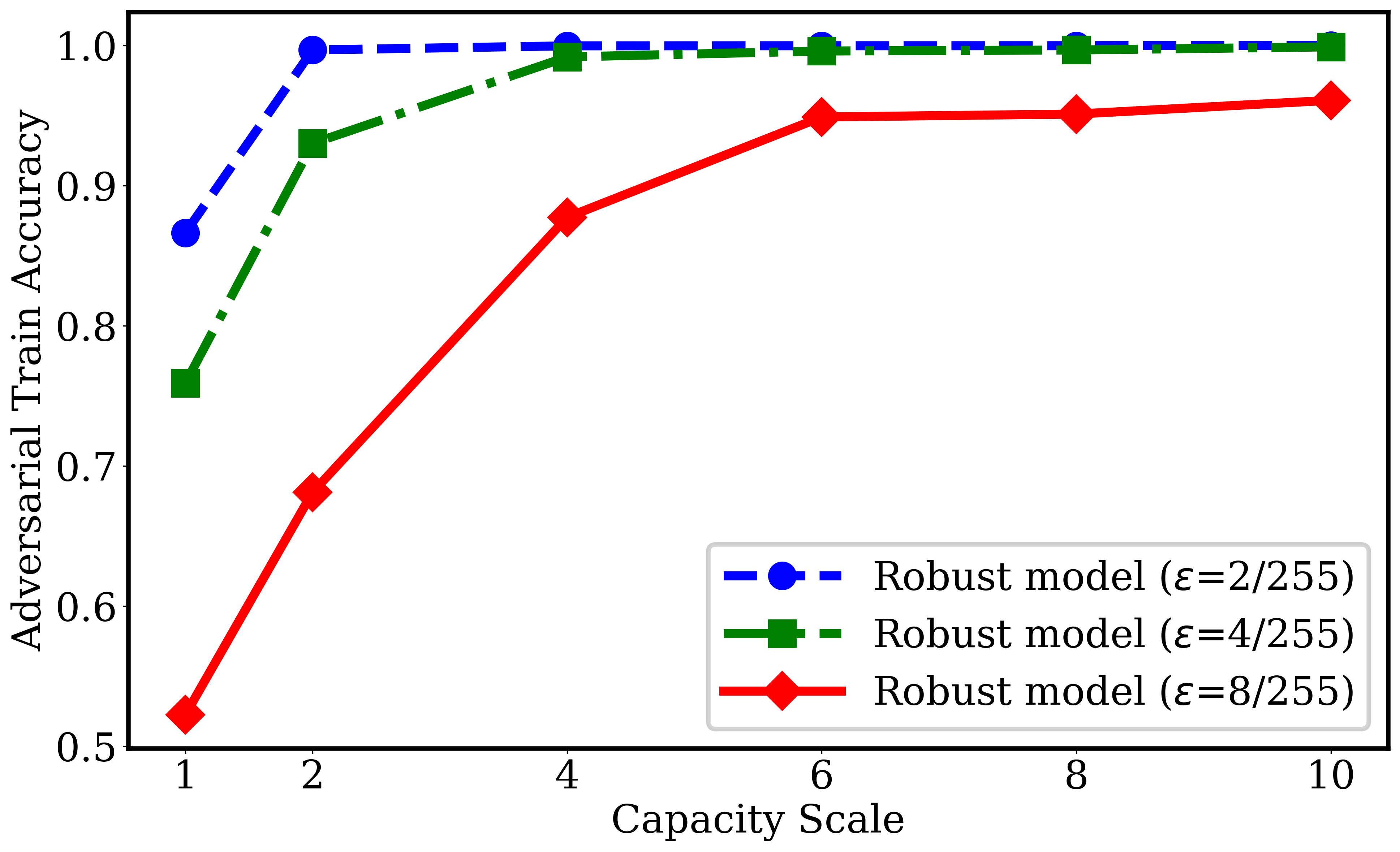}
		\caption{Adversarial train accuracy for models with different model capacities.}
		\label{fig:robust_capacity}
	\end{subfigure}
	\caption{Membership inference accuracy and adversarial train accuracy for CIFAR10 classifiers \cite{madry_robust_ICLR18} with varying model capacities. 
	The model with a capacity scale of $s$ contains 3 groups of residual layers with output channel numbers ($16s$, $32s$, $64s$), as described in Section \ref{sec:setup}.
	}
	\label{fig:adv_cifar_capacity}
\end{figure}

Here we investigate the influence of model capacity by varying the capacity scale of wide ResNet architecture \cite{Zagoruyko_WRN_BMVC16} used in CIFAR10 training, 
which is proportional to the output channel numbers of residual layers. 
We perform membership inference attacks for the robust models, and show the results in Figure \ref{fig:adv_cifar_capacity}. 
The attacks are based on benign inputs' predictions (strategy $\mathcal{I}_{\boldsymbol{\mathrm{B}}}$) and 
the gray line measures the privacy leakage for the natural models as a baseline.

First, we can see that \textbf{as the model capacity increases, the model has a higher membership inference accuracy}, along with a higher adversarial train accuracy. 
Second, \textbf{when using a larger adversarial perturbation budget $\epsilon$, a larger model capacity is also needed}. 
When $\epsilon=2/255$, a capacity scale of 2 is enough to fit the training data, while for $\epsilon=8/255$, a capacity scale of 8 is needed.

\subsubsection{Inference attacks using targeted adversarial examples}\label{subsubsec:targeted_adv}
Next, we investigate membership inference attacks using \emph{targeted} adversarial examples. 
For each input, we compute 9 targeted adversarial examples with each of the 9 incorrect labels as targets  using Equation \eqref{eq:PGD-target}. 
We then compute the output prediction vectors for all adversarial examples 
and use the shadow-training inference method proposed by Shokri et al. \cite{shokri_membership_SP17} to perform membership inference attacks.
Specifically, for each class label, we learn a dedicated inference model (binary classifier) 
by using the output predictions of targeted adversarial examples from $500$ training points and $500$ test points as the training set for the membership inference. 
We then test the inference model on the remaining CIFAR10 training and test examples from the same class label. 
In our experiments, we use a 3-layer fully connected neural network with size of hidden neurons equal to 200, 20, and 2 respectively. 
We call this method ``model-infer (targeted)''.

For untargeted adversarial examples or benign examples, a similar class label-dependent inference model can also be obtained 
by using either untargeted adversarial example's prediction vector or benign example's prediction vector as features of the inference model. 
We call these methods  ``model-infer (untargeted)'' and ``model-infer (benign)''. We use the same 3-layer fully connected neural network as the inference classifier.

Finally, we also adapt our confidence-thresholding inference strategy to be class-label dependent 
by choosing the confidence threshold value according to prediction confidence values from $500$  training points and $500$  test points, 
and then testing on remaining CIFAR10 points from the same class label. 
Based on whether the confidence value is from the untargeted adversarial input or the benign input, 
we call the method as ``confidence-infer (untargeted)'' and ``confidence-infer (benign)''.

\begin{table}[!htb]
\caption{Comparison of membership inference attacks against the robust CIFAR10 classifier \cite{madry_robust_ICLR18}. 
Inference attack strategies include combining predictions of targeted adversarial examples, untargeted adversarial examples, and benign examples 
with either training an inference neural network model or thresholding the prediction confidence.
}
\centering
\renewcommand\arraystretch{2}
\fontsize{6.5pt}{6.5pt}\selectfont
\begin{tabular}{|c|c|c|c|c|c|}
\hline
Class & confidence-infer & model-infer & confidence-infer & model-infer & model-infer \\
label & (benign) & (benign) & (untargeted) & (untargeted) & (targeted) \\
\hline
0 & 70.88\% & 71.49\%  & 72.21\% & 72.70\% & \textbf{74.42\% }  \\
\hline
1  & 63.57\%  & 64.42\% & 67.52\% & 67.69\% & \textbf{68.88\%}   \\
\hline
2 & 80.16\%  & 76.74\%  & 79.71\% &  80.16\% & \textbf{83.58\%}   \\
\hline
3 & 90.43\% & 90.49\%   & 87.64\% & 87.83\% & \textbf{90.57\% }  \\
\hline
4 & 82.30\%  & 82.17\%  & 81.83\% & 81.57\% & \textbf{84.47\% }  \\
\hline
5 & 81.34\% & 79.84\%   & 81.57\% & 81.34\% & \textbf{83.02\%}   \\
\hline
6 & 75.34\%  & 70.92\%  & 77.66\% & 76.97\% & \textbf{79.94\%}   \\
\hline
7 & 69.54\% & 67.61\%   & 72.92\% & 72.82\% & \textbf{72.98\% }  \\
\hline
8 & 69.16\% & 69.57\%   & 74.36\% & 74.40\% & \textbf{75.33\% }  \\
\hline
9 & 68.13\%  & 66.34\%  & 71.86\% & 72.06\% & \textbf{73.32\%}   \\
\hline
\end{tabular}
\label{tab:adv_cifar_inferences_compare}
\end{table}

The membership inference attack results using the above five strategies are presented in Table \ref{tab:adv_cifar_inferences_compare}. 
We can see that the \textbf{targeted adversarial example based inference strategy ``model-infer (targeted)'' always has the highest inference accuracy.}
This is because the targeted adversarial examples contain information about distance of the input to each label's decision boundary, 
while untargeted adversarial examples contain information about distance of the input to only a nearby label's decision boundary.
Thus targeted adversarial examples leak more membership information.
As an aside, we also find that our confidence-based inference methods obtain nearly the same inference results as training neural network models, 
showing the effectiveness of the confidence-thresholding inference strategies.

\begin{table*}[!htb]
\caption{Membership inference attacks against natural and verifiably robust models \cite{wong_robust_ICML18, mirman_provable_ICML18, gowal_IBP_secml18} on the Yale Face dataset with a $l_{\infty}$ perturbation constraint $\epsilon=8/255$. Based on Equation \eqref{eq:inf_advantage}, the natural model has the inference advantage of $11.70\%$, while the robust model has the inference advantage up to $52.10\%$.
}
\centering
\renewcommand\arraystretch{1.3}
\fontsize{6.7pt}{6.7pt}\selectfont
\begin{tabular}{|c|c|c|c|c|c|c|c|c|c|}
\hline
Training & train & test & adv-train & adv-test & ver-train & ver-test & inference & inference & inference \\
method & acc & acc & acc & acc & acc & acc & acc ($\mathcal{I}_{\boldsymbol{\mathrm{B}}}$) & acc ($\mathcal{I}_{\boldsymbol{\mathrm{A}}}$) & acc ($\mathcal{I}_{\boldsymbol{\mathrm{V}}}$) \\
\hline
\multirow{2}{*}{Natural} & \multirow{2}{*}{100\%} & \multirow{2}{*}{98.25\%}  & \multirow{2}{*}{4.53\%} & \multirow{2}{*}{2.92\%} & \multirow{2}{*}{N.A.} & \multirow{2}{*}{N.A.} & \multirow{2}{*}{\textbf{55.85\%}} & \multirow{2}{*}{54.27\%} & \multirow{2}{*}{N.A.}  \\
 & & & & & & & & & \\
\hline
{Dual-Based} & \multirow{2}{*}{98.89\%} & \multirow{2}{*}{92.80\%}  & \multirow{2}{*}{98.53\%} & \multirow{2}{*}{83.66\%} & \multirow{2}{*}{96.37\%} & \multirow{2}{*}{68.87\%} & \multirow{2}{*}{55.90\%} & \multirow{2}{*}{60.40\%} & \multirow{2}{*}{\textbf{64.48\%}} \\
Verify \cite{wong_robust_ICML18} & & & & & & & & & \\
\hline
Abs-Based & \multirow{2}{*}{99.26\%} & \multirow{2}{*}{83.27\%}  & \multirow{2}{*}{85.68\%} & \multirow{2}{*}{50.39\%} & \multirow{2}{*}{43.32\%} & \multirow{2}{*}{18.09\%} & \multirow{2}{*}{65.11\%} & \multirow{2}{*}{65.64\%} & \multirow{2}{*}{\textbf{67.05\%}} \\
Verify \cite{mirman_provable_ICML18} & & & & & & & & & \\
\hline
{IBP-Based}& \multirow{2}{*}{99.16\%} & \multirow{2}{*}{85.80\%}  & \multirow{2}{*}{94.42\%} & \multirow{2}{*}{69.68\%} & \multirow{2}{*}{89.58\%} & \multirow{2}{*}{36.77\%} & \multirow{2}{*}{60.45\%} & \multirow{2}{*}{66.28\%} & \multirow{2}{*}{\textbf{76.05\%}} \\ 
Verify \cite{gowal_IBP_secml18} & & & & & & & & & \\
\hline
\end{tabular}
\label{tab:verifiable_defenses_privacy_YALEFACE}
\end{table*}

\begin{table*}[!htb]
\caption{Membership inference attacks against natural and verifiably robust models \cite{wong_robust_ICML18, mirman_provable_ICML18, gowal_IBP_secml18} on the Fashion-MNIST dataset with a $l_{\infty}$ perturbation constraint $\epsilon=0.1$.
}
\centering
\renewcommand\arraystretch{1.3}
\fontsize{6.7pt}{6.7pt}\selectfont
\begin{tabular}{|c|c|c|c|c|c|c|c|c|c|}
\hline
Training & train & test & adv-train & adv-test & ver-train & ver-test & inference & inference & inference \\
method & acc & acc & acc & acc & acc & acc & acc ($\mathcal{I}_{\boldsymbol{\mathrm{B}}}$) & acc ($\mathcal{I}_{\boldsymbol{\mathrm{A}}}$) & acc ($\mathcal{I}_{\boldsymbol{\mathrm{V}}}$) \\
\hline
\multirow{2}{*}{Natural} & \multirow{2}{*}{100\%} & \multirow{2}{*}{92.18\%}  & \multirow{2}{*}{4.35\%} & \multirow{2}{*}{4.14\%} & \multirow{2}{*}{N.A.} & \multirow{2}{*}{N.A.} & \multirow{2}{*}{\textbf{57.12\%}} & \multirow{2}{*}{50.95\%} & \multirow{2}{*}{N.A.} \\
 & & & & & & & & & \\
\hline
{Dual-Based} & \multirow{2}{*}{75.13\%} & \multirow{2}{*}{74.29\%}  & \multirow{2}{*}{65.77\%} & \multirow{2}{*}{65.36\%} & \multirow{2}{*}{61.77\%} & \multirow{2}{*}{61.45\%} & \multirow{2}{*}{\textbf{50.58\%}} & \multirow{2}{*}{50.42\%} & \multirow{2}{*}{50.45\%} \\
Verify \cite{wong_robust_ICML18} & & & & & & & & & \\
\hline
{Abs-Based} & \multirow{2}{*}{86.44\%} & \multirow{2}{*}{85.47\%}  & \multirow{2}{*}{74.12\%} & \multirow{2}{*}{73.28\%} & \multirow{2}{*}{69.69\%} & \multirow{2}{*}{68.89\%} & \multirow{2}{*}{\textbf{50.79\%}} & \multirow{2}{*}{50.69\%} & \multirow{2}{*}{50.59\%} \\
Verify \cite{mirman_provable_ICML18} & & & & & & & & & \\
\hline
IBP-Based & \multirow{2}{*}{89.85\%} & \multirow{2}{*}{86.26\%}  & \multirow{2}{*}{82.60\%} & \multirow{2}{*}{78.44\%} & \multirow{2}{*}{79.20\%} & \multirow{2}{*}{74.17\%} & \multirow{2}{*}{52.13\%} & \multirow{2}{*}{52.06\%} & \multirow{2}{*}{\textbf{52.67\%}} \\ 
Verify \cite{gowal_IBP_secml18} & & & & & & & & & \\
\hline
\end{tabular}
\label{tab:verifiable_defenses_privacy_FMNIST}
\end{table*}

\begin{table*}[!htb]
\caption{Membership inference attacks against natural and verifiably robust CIFAR10 classifiers \cite{wong_robust_ICML18} 
trained on a subset ($20\%$) of the training data with varying $l_{\infty}$ perturbation budgets.
}
\centering
\renewcommand\arraystretch{1.3}
\fontsize{6.7pt}{6.7pt}\selectfont
\begin{tabular}{|c|c|c|c|c|c|c|c|c|c|c|}
\hline
Training & Perturbation & train & test & adv-train & adv-test & ver-train & ver-test & inference & inference & inference \\
method & budgets ($\epsilon$) & acc & acc & acc & acc & acc & acc & acc ($\mathcal{I}_{\boldsymbol{\mathrm{B}}}$) & acc ($\mathcal{I}_{\boldsymbol{\mathrm{A}}}$) & acc ($\mathcal{I}_{\boldsymbol{\mathrm{V}}}$) \\
\hline
\multirow{2}{*}{Natural} & \multirow{2}{*}{N.A.} & \multirow{2}{*}{99.83\%} & \multirow{2}{*}{71.80\%}  & \multirow{2}{*}{N.A.} & \multirow{2}{*}{N.A.} & \multirow{2}{*}{N.A.} & \multirow{2}{*}{N.A.} & \multirow{2}{*}{\textbf{71.50\%}} & \multirow{2}{*}{N.A.} & \multirow{2}{*}{N.A.} \\
& & & & & & & & & & \\
\hline
{Dual-Based} & \multirow{2}{*}{0.25/255} & \multirow{2}{*}{100\%} & \multirow{2}{*}{73.10\%}  & \multirow{2}{*}{99.99\%} & \multirow{2}{*}{69.84\%} & \multirow{2}{*}{99.99\%} & \multirow{2}{*}{68.18\%} & \multirow{2}{*}{76.13\%} & \multirow{2}{*}{\textbf{76.18\%}} & \multirow{2}{*}{76.04\%} \\
Verify \cite{wong_robust_ICML18} & & & & & & & & & & \\
\hline
{Dual-Based} & \multirow{2}{*}{0.5/255} & \multirow{2}{*}{99.98\%} & \multirow{2}{*}{69.29\%}  & \multirow{2}{*}{99.98\%} & \multirow{2}{*}{64.51\%} & \multirow{2}{*}{99.97\%} & \multirow{2}{*}{60.89\%} & \multirow{2}{*}{77.06\%} & \multirow{2}{*}{\textbf{77.36\%}} & \multirow{2}{*}{77.09\%} \\ 
Verify \cite{wong_robust_ICML18} & & & & & & & & & & \\
\hline
{Dual-Based} & \multirow{2}{*}{0.75/255} & \multirow{2}{*}{100\%} & \multirow{2}{*}{65.25\%}  & \multirow{2}{*}{99.95\%} & \multirow{2}{*}{59.49\%} & \multirow{2}{*}{99.85\%} & \multirow{2}{*}{54.71\%} & \multirow{2}{*}{77.99\%} & \multirow{2}{*}{\textbf{78.50\%}} & \multirow{2}{*}{78.20\%} \\
Verify \cite{wong_robust_ICML18} & & & & & & & & & & \\
\hline
{Dual-Based} &\multirow{2}{*}{1/255} & \multirow{2}{*}{99.78\%} & \multirow{2}{*}{63.96\%}  & \multirow{2}{*}{99.44\%} & \multirow{2}{*}{57.06\%} & \multirow{2}{*}{98.61\%} & \multirow{2}{*}{50.74\%} & \multirow{2}{*}{76.30\%} & \multirow{2}{*}{77.05\%} & \multirow{2}{*}{\textbf{77.16\%}} \\
Verify \cite{wong_robust_ICML18} & & & & & & & & & & \\
\hline
{Dual-Based} & \multirow{2}{*}{1.25/255} & \multirow{2}{*}{98.46\%} & \multirow{2}{*}{61.79\%}  & \multirow{2}{*}{97.30\%} & \multirow{2}{*}{53.76\%} & \multirow{2}{*}{95.36\%} & \multirow{2}{*}{46.70\%} & \multirow{2}{*}{74.07\%} & \multirow{2}{*}{75.10\%} & \multirow{2}{*}{\textbf{75.41\%}} \\
Verify \cite{wong_robust_ICML18} & & & & & & & & & & \\
\hline
{Dual-Based} & \multirow{2}{*}{1.5/255} & \multirow{2}{*}{96.33\%} & \multirow{2}{*}{60.97\%}  & \multirow{2}{*}{94.27\%} & \multirow{2}{*}{51.72\%} & \multirow{2}{*}{90.19\%} & \multirow{2}{*}{44.23\%} & \multirow{2}{*}{71.08\%} & \multirow{2}{*}{72.29\%} & \multirow{2}{*}{\textbf{72.69\%}} \\
Verify \cite{wong_robust_ICML18} & & & & & & & & & & \\
\hline
\end{tabular}
\label{tab:dual_relax_partial_cifar_privacy}
\end{table*}

\section{Membership Inference Attacks against Verifiably Robust Models}\label{sec:results_verifiably_robust}

In this section we perform membership inference attacks against 3 verifiable defense methods: duality-based verification (Dual-Based Verify) \cite{wong_robust_ICML18}, 
abstract interpretation-based verification (Abs-Based Verify) \cite{mirman_provable_ICML18}, 
and interval bound propagation-based verification (IBP-Based Verify) \cite{gowal_IBP_secml18}.
We train the verifiably robust models using the network architectures as described in Section \ref{sec:setup} 
(with minor modifications for the Dual-Based Verify method \cite{wong_robust_ICML18} as discussed in Appendix \ref{sec:append_changes}), 
the $l_{\infty}$ perturbation budget $\epsilon$ is set to be $8/255$ for the Yale Face dataset and $0.1$ for the Fashion-MNIST dataset.
We do not evaluate the verifiably robust models for the full CIFAR10 dataset as none of these three defense methods scale to the wide ResNet architecture.

\subsection{Overall Results}

The membership inference attack results against natural and verifiably robust models are presented in Table \ref{tab:verifiable_defenses_privacy_YALEFACE} 
and Table \ref{tab:verifiable_defenses_privacy_FMNIST}, 
where ``acc'' stands for accuracy, ``adv-train acc'' and ``adv-test acc'' measure adversarial accuracy under PGD attacks (Equation \eqref{eq:PGD}), 
and ``ver-train acc'' and ``ver-test acc'' report the verified worse-case accuracy under the perturbation constraint $\mathcal{B}_{\epsilon}$.

\textbf{For the Yale Face dataset, all three defense methods leak more membership information}.
The IBP-Based Verify method even leads to an inference accuracy above $75\%$, 
higher than the inference accuracy of empirical defenses shown in Table \ref{tab:empirical_defenses_privacy_YALEFACE}, 
resulting a $4.5 \times$ membership inference advantage (Equation \eqref{eq:inf_advantage}) than the natural model.
The inference strategy based on verified prediction confidence (strategy $\mathcal{I}_{\boldsymbol{\mathrm{V}}}$) has the highest inference accuracy 
as the verification process enlarges prediction confidence between training data and test data.

On the other hand, for the Fashion-MNIST dataset, we fail to obtain increased membership inference accuracies on the verifiably robust models.
However, we also observe much reduced benign train accuracy (below 90\%) and verified train accuracy (below 80\%), which means that 
\textbf{the model fits the training set poorly}.
Similar to our analysis of empirical defenses, we can think the verifiable defense as 
adding more ``virtual training points'' around each training example to compute its verified robust loss. 
Since the verified robust loss is an upper bound on the real robust loss, the added ``virtual training points'' are in fact beyond the $l_{\infty}$ ball. 
Therefore, the model capacity needed for verifiable defenses is even larger than that of empirical defense methods.

From the experiment results in Section \ref{subsubsec:PGD_defense_capacity}, we have shown that if the model capacity is not large enough, 
the robust model will not fit the training data well. 
This explains why membership inference accuracies for verifiably robust models are limited in Table \ref{tab:verifiable_defenses_privacy_FMNIST}.
However, enlarging the model capacity does not guarantee that the training points will fit well for verifiable defenses 
because the verified upper bound of robust loss is likely to be looser with a deeper and larger neural network architecture.
We validate our hypothesis in the following two subsections.

\subsection{Varying Model Capacities}\label{subsec:capacity_verifiable_defenses}

\begin{figure}[!ht]
	\centering
	\includegraphics[width=\linewidth]{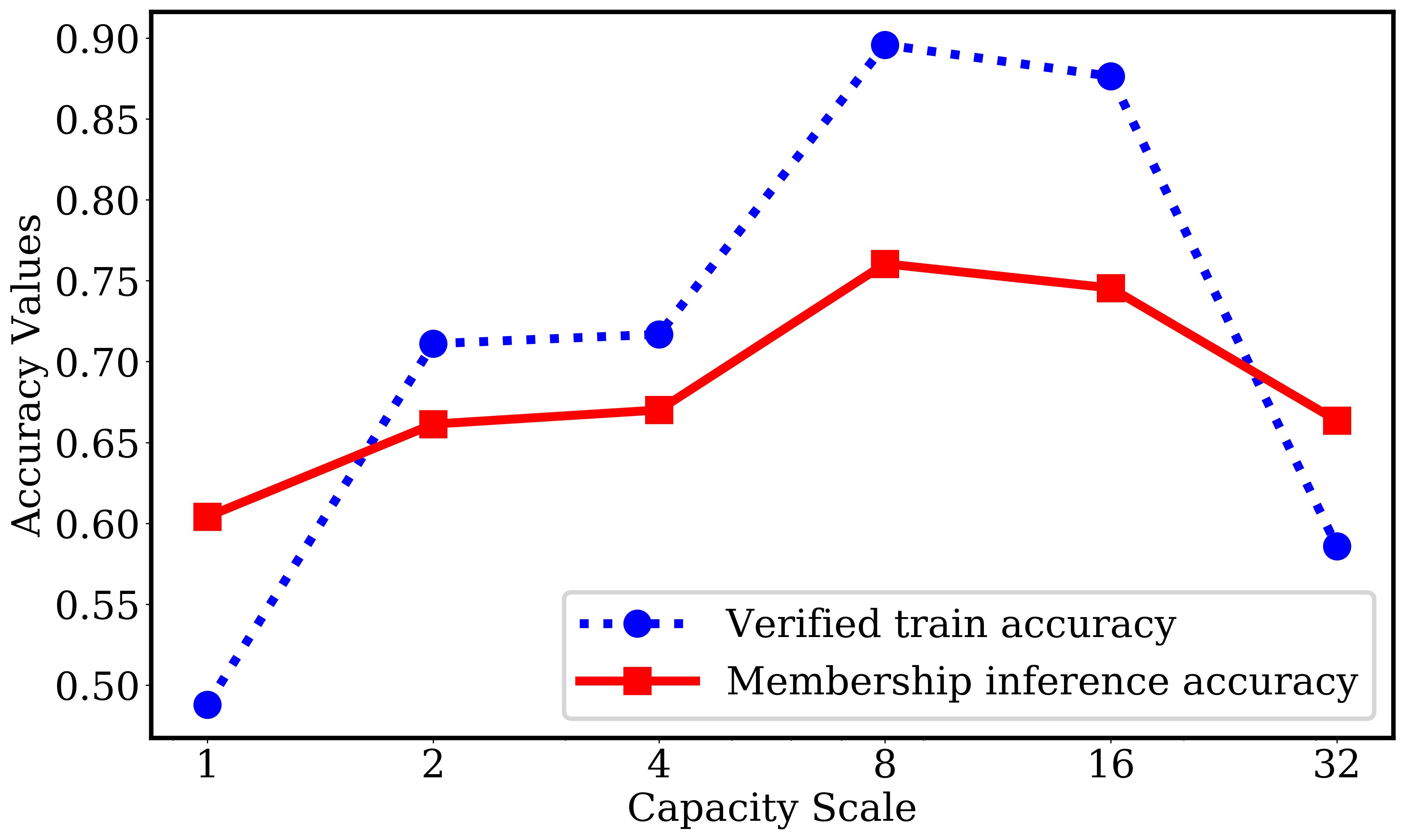}
	\caption{Verified train accuracy and membership inference accuracy using inference strategy $\mathcal{I}_{\boldsymbol{\mathrm{V}}}$ for 
	robust Yale Face classifiers \cite{gowal_IBP_secml18} with varying capacities.
	The model with a capacity scale of $s$ contains 4 convolution blocks with output channel numbers $(s, 2s, 4s, 8s)$, as described in Section \ref{sec:setup}.}.
	\label{fig:capacity_varifiable_defenses}
\end{figure}

We use models with varying capacities to robustly train on the Yale Face dataset with the IBP-Based Verify defense \cite{gowal_IBP_secml18} as an example.

We present the results in Figure \ref{fig:capacity_varifiable_defenses}, where model capacity scale of $8$ corresponds to the original model architecture, and we perform 
membership inference attacks based on verified worst-case prediction confidence $\mathcal{I}_{\boldsymbol{\mathrm{V}}}$.
We can see that when model capacity increases, at the beginning, robustness performance gets improved, 
and we also have a higher membership inference accuracy. 
However, when the model capacity is too large, the robustness performance and the membership inference accuracy begin decreasing, 
since now the verified robust loss becomes too loose.
\subsection{Reducing the Size of Training Set}

In this subsection, we further prove our hypothesis by showing that when the size of the training  set is reduced so that the model can fit well on the reduced dataset, 
the verifiable defense method indeed leads to an increased membership inference accuracy.

We choose the duality-based verifiable defense method \cite{wong_robust_ICML18, wong_robust_NIPS18} and train the CIFAR10 classifier with a normal ResNet architecture: 
3 groups of residual layers with output channel numbers (16, 32, 64) and only 1 residual unit for each group.
The whole CIFAR10 training set have too many points to be robustly fitted with the verifiable defense algorithm: 
the robust CIFAR10 classifier \cite{wong_robust_NIPS18} with $\epsilon=2/255$ has the train accuracy below $70\%$.
Therefore, we select a subset of the training data to robustly train the model by randomly choosing $1000$ ($20\%$) training images for each class label.
We vary the perturbation budget value ($\epsilon$) in order to observe when the model capacity is not large enough to fit on this partial CIFAR10 set 
using the verifiable training algorithm \cite{wong_robust_ICML18}.

We show the obtained results in Table \ref{tab:dual_relax_partial_cifar_privacy}, where the natural model has a low test accuracy (below 75\%) 
and high privacy leakage (inference accuracy is $71.50\%$) since we only use $20\%$ training examples to learn the classifier.
By using the verifiable defense method \cite{wong_robust_ICML18}, 
\textbf{the verifiably robust models have increased membership inference accuracy values}, for all $\epsilon$ values.
We can also see that when increasing the $\epsilon$ values, at the beginning, the robust model is more and more susceptible to membership inference attacks 
(inference accuracy increases from $71.50\%$ to $78.50\%$). 
However, beyond a threshold of $\epsilon=1/255$, the inference accuracy starts to decrease, 
since a higher $\epsilon$ requires a model with a larger capacity to fit well on the training data.

\section{Discussions}

In this section, we first evaluate the success of membership inference attacks when the adversary does not know the $l_{\infty}$ perturbation constraints of robust models.
Second we discuss potential countermeasures, including temperature scaling and regularization, to reduce privacy risks.
Finally, we discuss the relationship between training data privacy and model robustness.

\subsection{Membership Inference Attacks with Varying Perturbation Constraints}\label{subsec:different-lp-balls}

Our experiments so far considered an adversary with prior knowledge of the robust model's $l_{\infty}$ perturbation constraint.
Next, we evaluate privacy leakage of robust models in the absence of such prior knowledge by varying perturbation budgets used in the membership inference attack.
Specifically, we perform membership inference attacks $\mathcal{I}_{\boldsymbol{\mathrm{A}}}$ with varying perturbation constraints against robust Yale Face classifiers \cite{madry_robust_ICLR18, sinha_PAT_ICLR18, zhang_TRADES_ICML19}, which are robustly trained with the $l_{\infty}$ perturbation budget ($\epsilon$) of $8/255$.

\begin{figure}[!ht]
	\centering
	\includegraphics[width=\linewidth]{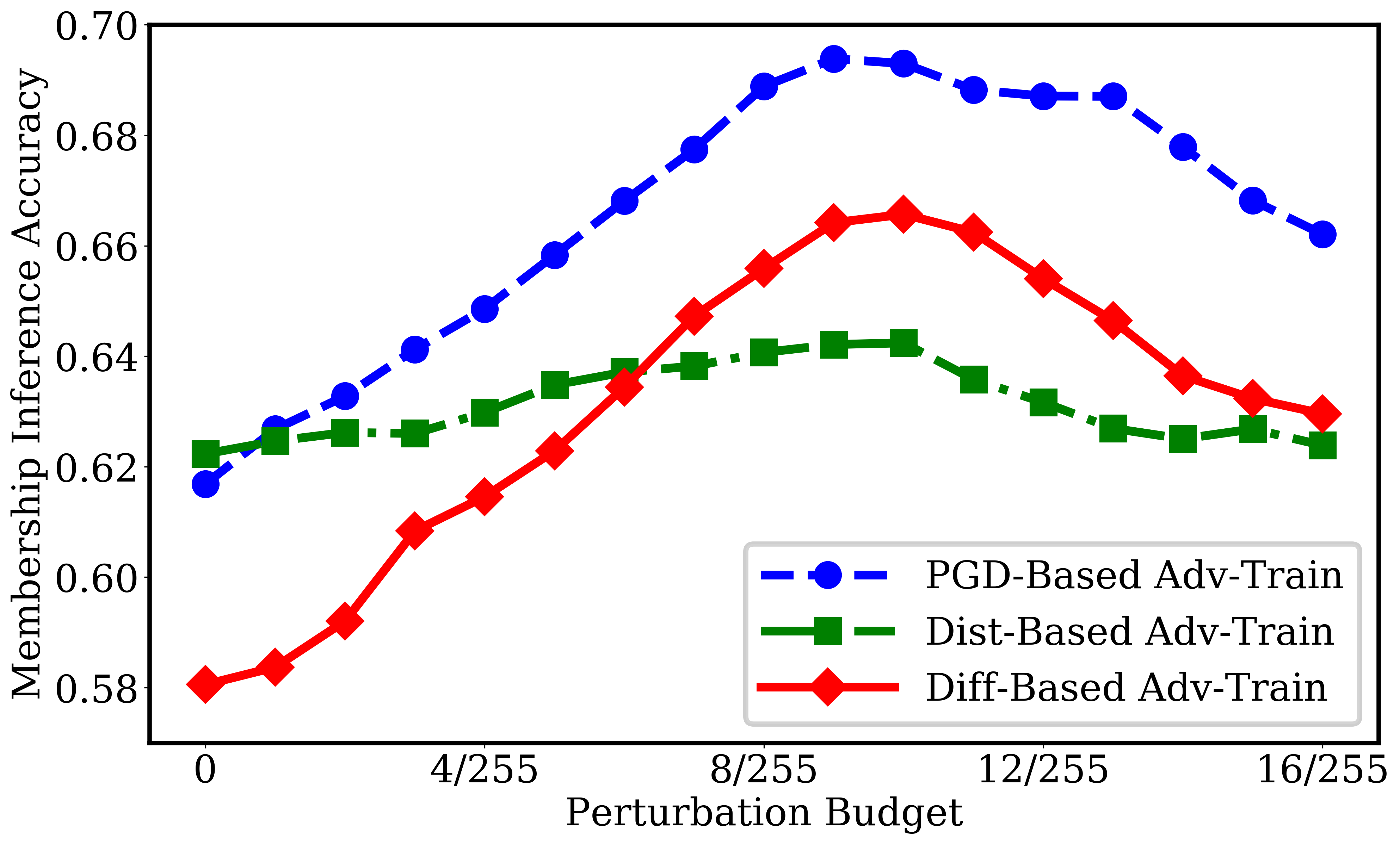}
	\caption{Membership inference accuracy on robust Yale Face classifiers \cite{madry_robust_ICLR18, sinha_PAT_ICLR18, zhang_TRADES_ICML19} trained with the $l_{\infty}$ perturbation constraint of $8/255$.
	The privacy leakage is evaluated via the inference strategy $\mathcal{I}_{\boldsymbol{\mathrm{A}}}$ based on adversarial examples generated with varying perturbation budgets.
	}
	\label{fig:yale_varied_eps}
\end{figure}

We present the membership inference results in Figure \ref{fig:yale_varied_eps}, where the inference strategy $\mathcal{I}_{\boldsymbol{\mathrm{A}}}$ with the perturbation budget of $0$ is equivalent to the inference strategy $\mathcal{I}_{\boldsymbol{\mathrm{B}}}$.
In general, we observe a higher membership inference accuracy when the perturbation budget used in the inference attack is close to the robust model's exact perturbation constraint. 
An attack perturbation budget that is very small will not fully utilize the classifier's structural characteristics, leading to high robustness performance for adversarial examples generated from both training and test data.
On the other hand, a very large perturbation budget leads to low accuracy on adversarial examples generated from both training training data and test data. 
Both of these scenarios will reduce the success of membership inference attacks.

Based on results shown in Figure \ref{fig:yale_varied_eps}, the adversary does not need to know the exact value of robust model's $l_{\infty}$ perturbation budget: approximate knowledge of $\epsilon$ suffices to achieve high membership inference accuracy.
Furthermore, the adversary can leverage the shadow training technique (with shadow training set) \cite{shokri_membership_SP17} in practice to compute the best attack parameters (the perturbation budget and the threshold value), and then use the inferred parameters against the target model. The best perturbation budget may not even be same as the exact $\epsilon$ value of robust model.
For example, we obtain the highest membership inference accuracy by setting $\epsilon$ as $9/255$ for the PGD-Based Adv-Train Yale Face classifier \cite{madry_robust_ICLR18}, and $10/255$ for the other two robust classifiers \cite{sinha_PAT_ICLR18, zhang_TRADES_ICML19}.
We observe similar results for Fashion-MNIST and CIFAR10 datasets, which are presented in Appendix \ref{sec:different_eps_fmnist&cifar}.

\subsection{Potential Countermeasures}

We discuss potential countermeasures that can reduce the risk of membership inference attacks while maintaining model robustness. 

\subsubsection{Temperature scaling}
Our membership inference strategies leverage the difference between the prediction confidence of the target model on its training set and test set.
Thus, a straightforward mitigation method is to reduce this difference by applying temperature scaling on logits \cite{guo_temperature_ICML17}.
The temperature scaling method was shown to be effective to reduce privacy risk for natural (baseline) models by Shokri et al. \cite{shokri_membership_SP17}, 
while we are studying its effect for robust models here.

Temperature scaling is a post-processing calibration technique for machine learning models that divides logits by the temperature, $T$, before the softmax function. 
Now the model prediction probability can be expressed as
\begin{equation}\label{eq:temperature_scale}
F(\bfx)_{i} = \frac{\exp \,(g(\bfx)_{i}/T)}{\sum_{j=0}^{k-1} \exp \,(g(\bfx)_{j}/T)},
\end{equation}
where $T=1$ corresponds to original model prediction.
By setting $T>1$, the prediction confidence $F(\bfx)_{y}$ is reduced, 
and when $T \rightarrow \infty$, the prediction output is close to uniform and independent of the input, 
thus leaking no membership information while making the model useless for prediction.

\begin{figure}[!ht]
	\centering
	\includegraphics[width=\linewidth]{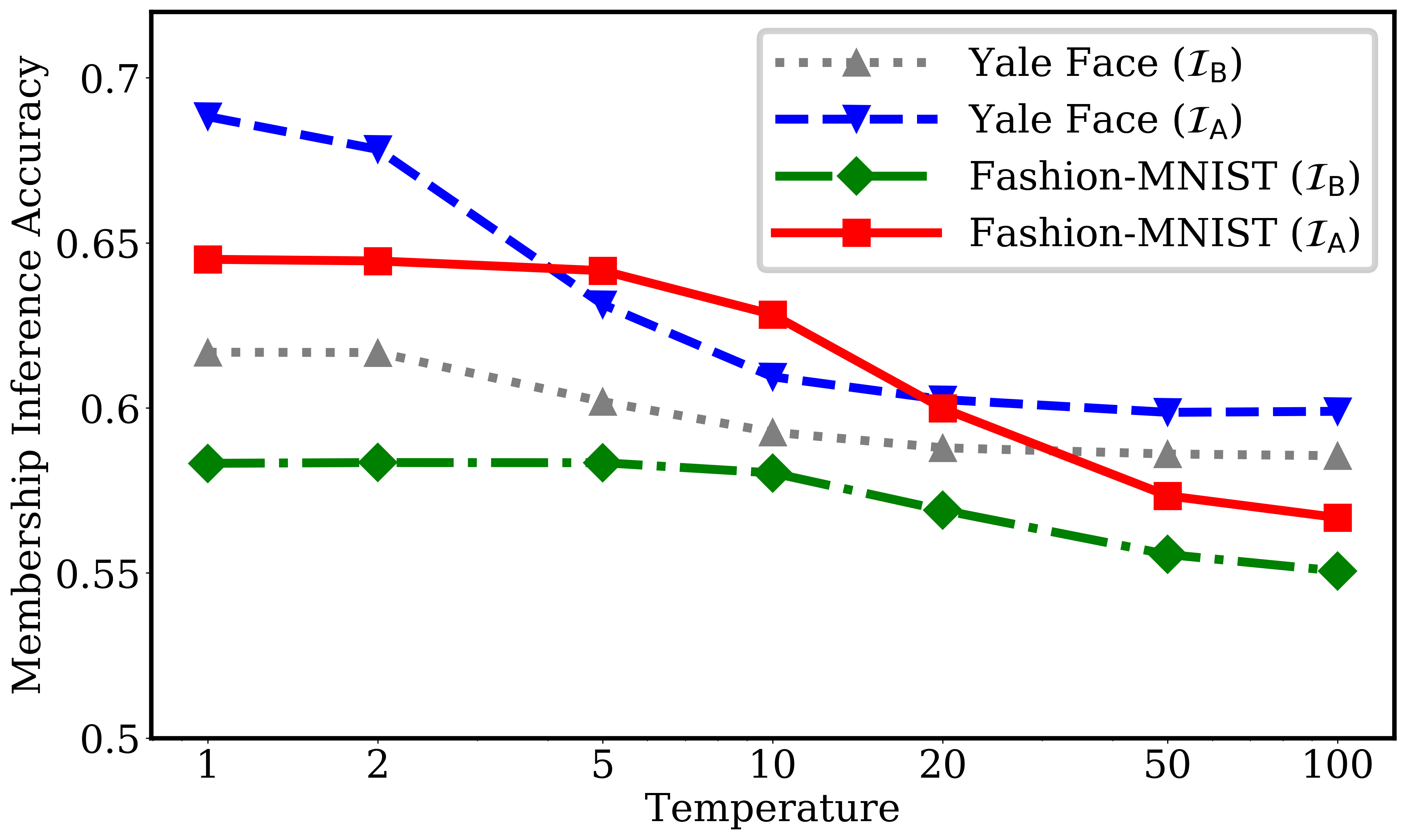}
	\caption{Membership inference accuracy on robust Yale Face and Fashion-MNIST classifiers \cite{madry_robust_ICLR18} with varying softmax temperature values \cite{guo_temperature_ICML17}.
	}
	\label{fig:temperature_scale}
\end{figure}

We apply the temperature scaling technique on the robust Yale Face and Fashion-MNIST classifiers 
using the PGD-based adversarial training defense \cite{madry_robust_ICLR18} and investigate its effect on membership inference.
We present membership inference results (both $\mathcal{I}_{\boldsymbol{\mathrm{B}}}$ and $\mathcal{I}_{\boldsymbol{\mathrm{A}}}$) for varying temperature values (while maintaining the same classification accuracy) in Figure \ref{fig:temperature_scale}. 
We can see that increasing the temperature value decreases the membership inference accuracy.

\subsubsection{Regularization to improve robustness generalization}
\label{subsub:robustness_regularization}
Regularization techniques such as parameter norm penalties and dropout \cite{srivastava_dropout14}, 
are typically used during the training process to solve overfitting issues for machine learning models.
Shokri et al. \cite{shokri_membership_SP17} and Salem et al. \cite{salem_membership_NDSS19} validate their effectiveness against membership inference attacks.
Furthermore, Nasr et al. \cite{nasr_membership_defense_CCS18} propose to measure the performance of membership inference attack at each training step 
and use the measurement as a new regularizer.

The above mitigation strategies are effective regardless of natural or robust machine learning models.
For the robust models, we can also rely on the regularization approach, which improves the model's robustness generalization. 
This can mitigate membership inference attacks, since a poor robustness generalization leads to a severe privacy risk.
We study the method proposed by Song et al. \cite{song_ATDA_ICLR19} to improve model's robustness generalization 
and explore its performance against membership inference attacks.

The regularization method in \cite{song_ATDA_ICLR19} performs domain adaptation (DA) \cite{torralba_DA_CVPR11} 
for the benign examples and adversarial examples on the logits: 
two multivariate Gaussian distributions for the logits of benign examples and adversarial examples are computed, 
and $l_{1}$ distances between two mean vectors and two covariance matrices are added into the training loss.

\begin{table}[!htb]
\caption{Membership inference attacks against robust models \cite{madry_robust_ICLR18}, 
where the perturbation budget $\epsilon$ is $8/255$ for the Yale Face datset, and $0.1$ for the Fashion-MNIST dataset.
When using DA, we modify the robust training algorithm by adding the regularization loss proposed by Song et al. \cite{song_ATDA_ICLR19}.
}
\centering
\renewcommand\arraystretch{1.3}
\fontsize{6.2pt}{6.2pt}\selectfont
\begin{tabular}{|c|c|c|c|c|c|c|c|}
\hline
\multirow{2}{*}{Dataset} &  using &  train & test & adv-train & adv-test & inference & inference \\
 &  DA \cite{song_ATDA_ICLR19}? & acc & acc & acc & acc & acc ($\mathcal{I}_{\boldsymbol{\mathrm{B}}}$) & acc ($\mathcal{I}_{\boldsymbol{\mathrm{A}}}$) \\
\hline
\multirow{2}{*}{Yale Face} & \multirow{2}{*}{no} & \multirow{2}{*}{99.89\%} & \multirow{2}{*}{96.69\%}  & \multirow{2}{*}{99.00\%} 
& \multirow{2}{*}{77.63\%} & \multirow{2}{*}{61.69\%} & \multirow{2}{*}{\textbf{68.83\%}} \\ & & & & & & &  \\
\hline
\multirow{2}{*}{Yale Face} & \multirow{2}{*}{yes} & \multirow{2}{*}{99.32\%} & \multirow{2}{*}{94.75\%}  & \multirow{2}{*}{99.26\%} 
& \multirow{2}{*}{88.52\%} & \multirow{2}{*}{60.73\%} & \multirow{2}{*}{\textbf{63.14\%}} \\ & & & & & & &  \\
\hline
Fashion & \multirow{2}{*}{no} & \multirow{2}{*}{99.93\%} & \multirow{2}{*}{90.88\%}  & \multirow{2}{*}{96.91\%} & \multirow{2}{*}{68.06\%} 
& \multirow{2}{*}{58.32\%} & \multirow{2}{*}{\textbf{64.49\%}} \\ MNIST & & & & & & &  \\
\hline
Fashion & \multirow{2}{*}{yes} & \multirow{2}{*}{88.97\%} & \multirow{2}{*}{86.98\%}  & \multirow{2}{*}{81.59\%} & \multirow{2}{*}{78.65\%} 
& \multirow{2}{*}{51.19\%} & \multirow{2}{*}{\textbf{51.49\%}} \\ MNIST & & & & & & &  \\
\hline
\end{tabular}
\label{tab:defense_by_ATDA}
\end{table}

We apply this DA-based regularization approach on the PGD-based adversarial training defense \cite{madry_robust_ICLR18} 
to investigate its effectiveness against membership inference attacks.
We list the experimental results both with and without the use of DA regularization for Yale Face and Fashion-MNIST datasets in Table \ref{tab:defense_by_ATDA}. 
We can see that the DA-based regularization can decrease the gap between adversarial train accuracy and adversarial test accuracy (robust generalization error), 
leading to a reduction in membership inference risk.

\subsection{Privacy vs Robustness}

We have shown that there exists a conflict between privacy of training data and model robustness:
all six robust training algorithms that we tested increase models' robustness against adversarial examples, but also make them more susceptible to membership inference attacks, compared with the natural training algorithm.
Here, we provide further insights on how general this relationship between membership inference and adversarial robustness is.

\subsubsection{Beyond image classification}

Our experimental evaluation so far  focused on the image classification domain. 
Next, we evaluate the privacy leakage of a robust model in a domain different than image classification to observe whether the conflict between privacy and robustness still holds.

We choose the UCI Human Activity Recognition (HAR) dataset \cite{anguita2013har}, which contains measurements of a smartphone's accelerometer and gyroscope values while the participants holding it performed one of six activities (walking, walking upstairs, walking downstairs, sitting, standing, and laying).
The dataset has 7,352 training samples and 2,947 test samples. Each sample is a $561$-feature vector with time and frequency domain variables of smartphone sensor values, and all features are normalized and bounded within [-1,1].

\begin{table}[!htb]
\caption{Membership inference attacks against natural and empirically robust models \cite{madry_robust_ICLR18} 
on the HAR dataset with a $l_{\infty}$ perturbation constraint $\epsilon=0.05$. 
Based on Equation \eqref{eq:inf_advantage}, the natural model has an inference advantage of $10.72\%$, while the robust model has an inference advantage of $20.26\%$.
}
\centering
\renewcommand\arraystretch{1.3}
\fontsize{6.7pt}{6.7pt}\selectfont
\begin{tabular}{|c|c|c|c|c|c|c|}
\hline
Training & train & test & adv-train & adv-test & inference & inference \\
method & acc & acc & acc & acc & acc ($\mathcal{I}_{\boldsymbol{\mathrm{B}}}$) & acc ($\mathcal{I}_{\boldsymbol{\mathrm{A}}}$) \\
\hline
\multirow{2}{*}{Natural} & \multirow{2}{*}{100\%} & \multirow{2}{*}{96.61\%}  & \multirow{2}{*}{33.56\%} & \multirow{2}{*}{29.69\%} & \multirow{2}{*}{\textbf{55.36\%}} & \multirow{2}{*}{55.03\%}  \\
 & & & & & & \\
\hline
PGD-Based & \multirow{2}{*}{96.10\%} & \multirow{2}{*}{92.53\%}  & \multirow{2}{*}{92.51\%} & \multirow{2}{*}{73.84\%} & \multirow{2}{*}{58.29\%} & \multirow{2}{*}{\textbf{60.13\%}} \\
Adv-Train \cite{madry_robust_ICLR18} & & & & & &  \\
\hline
\end{tabular}
\label{tab:har}
\end{table}

To train the classifiers, we use a 3-layer fully connected neural network with 1,000, 100, and 6 neurons respectively.
For robust training, we follow Wong and Kolter \cite{wong_robust_ICML18} by using the $l_{\infty}$ perturbation constraint with the size of $0.05$, and apply the PGD-based adversarial training \cite{madry_robust_ICLR18}.
The results for membership inference attacks against the robust classifier and its naturally trained counterpart are presented in Table \ref{tab:har}.
We can see that the robust training algorithm still leaks more membership information: the robust model has a $2 \times$ membership inference advantage (Equation \eqref{eq:inf_advantage}) over the natural model.

\subsubsection{Is the conflict a fundamental principle?}

It is difficult to judge whether the privacy-robustness conflict is \emph{fundamental} or not: will a robust training algorithm inevitably increase the model risk against membership inference attacks, compared to the natural training algorithm?
On the one hand, there is no direct tension between privacy of training data and model robustness.
We have shown in Section \ref{subsubsec:privacy_generalization} that the privacy leakage of robust model is related to its generalization error in the adversarial setting. 
The regularization method in Section \ref{subsub:robustness_regularization}, which improves the adversarial test accuracy and decreases the generalization error, indeed helps to decrease the membership inference accuracy.

On the other hand, our analysis verifies that state-of-the-art robust training algorithms \cite{madry_robust_ICLR18, sinha_PAT_ICLR18, zhang_TRADES_ICML19, wong_robust_ICML18, mirman_provable_ICML18, gowal_IBP_secml18} magnify the influence of training data on the model by minimizing the loss over a $l_p$ ball of each training point,
leading to more training data memorization. In addition, we find that a recently-proposed robust training algorithm \cite{lecuyer_PixelDP_SP19}, which adds a noise layer for robustness, also leads to an increase of membership inference accuracy in Appendix \ref{sec:other_robust_training}.
These robust training algorithms do not achieve good generalization of robustness performance \cite{schmidt_adv_trained_NIPS18, song_ATDA_ICLR19}. For example, even the regularized Yale Face classifier in Table \ref{tab:defense_by_ATDA} has a generalization error of $11\%$ in the adversarial setting, resulting a $2.3 \times$ membership inference advantage than the natural Yale Face classifier in Table \ref{tab:empirical_defenses_privacy_YALEFACE}.

Furthermore, the failure of  robustness generalization may partly be due to inappropriate (toy) distance constraints that are used to model adversaries.
Although $l_p$ perturbation constraints have been widely adopted in both attacks and defenses for adversarial examples \cite{biggio_evasion_KDD13, Goodfellow_evasion_arxiv14, madry_robust_ICLR18, wong_robust_ICML18}, the $l_p$ distance metric has limitations.
Sharif et al. \cite{sharif2018suitability} empirically show that (a) two images that are perceptually similar to humans can have a large $l_p$ distance, and (b) two images with a small $l_p$ distance can have different semantics.
Jacobsen et al. \cite{jacobsen2019exploiting} further show that robust training with a $l_p$ perturbation constraint makes the model more vulnerable to another type of adversarial examples: invariance based attacks that change the semantics of the image but leave the model predictions unchanged. 
Meaningful perturbation constraints to capture evasion attacks continue to be an important research challenge. 
We leave the question of deciding whether the privacy-robustness conflict is fundamental (i.e., will hold for next generation of defenses against adversarial examples) as an open question for the research community.
\section{Conclusions}

In this paper, we have connected both the security domain and the privacy domain for machine learning systems 
by investigating the membership inference privacy risk of robust training approaches (that mitigate the adversarial examples).
To evaluate the membership inference risk, we propose \emph{two new inference methods that exploit structural properties of adversarially robust defenses}, 
beyond the conventional inference method based on the prediction confidence of benign input.
By measuring the success of membership inference attacks on robust models trained with six state-of-the-art adversarial defense approaches, 
we find that \emph{all six robust training methods will make the machine learning model more susceptible to membership inference attacks, 
compared to the naturally undefended training}.
Our analysis further reveals that the privacy leakage is related to target model's robustness generalization, its adversarial perturbation constraint, and its capacity.
We also provide thorough discussions on the adversary's prior knowledge, potential countermeasures and the relationship between privacy and robustness. 
The detailed analysis in our paper highlights the importance of thinking about security and privacy together. 
Specifically, the membership inference risk needs to be considered when designing approaches to defend against adversarial examples.

\begin{acks}
We are grateful to anonymous reviewers at ACM CCS for valuable insights, and would like to specially thank Nicolas Papernot for shepherding the paper.
This work was supported in part by the National Science Foundation under grants CNS-1553437, CNS-1704105, CIF-1617286 and EARS-1642962,
by the Office of Naval Research Young Investigator Award,
by the Army Research Office Young Investigator Prize,
by Faculty research awards from Intel and IBM,
and by the National Research Foundation,
Prime Minister's Office, Singapore, under its Strategic Capability
Research Centres Funding Initiative.
\end{acks}

\bibliographystyle{ACM-Reference-Format}
\balance
\bibliography{refers}
\appendix
\section{Fine-Grained Analysis of Prediction Loss of the Robust CIFAR10 Classifier} \label{sec:fine_grained_analysis}

Here, we perform a fine-grained analysis of Figure \ref{fig:loss_adv_cifar} by separately visualizing the prediction loss distributions for test points 
which are secure and test points which are insecure. A point is deemed as secure when it is correctly classified by the model 
for all adversarial perturbations within the constraint $\mathcal{B}_{\epsilon}$.

\begin{figure}[!ht]
	\centering
	\includegraphics[width=\linewidth]{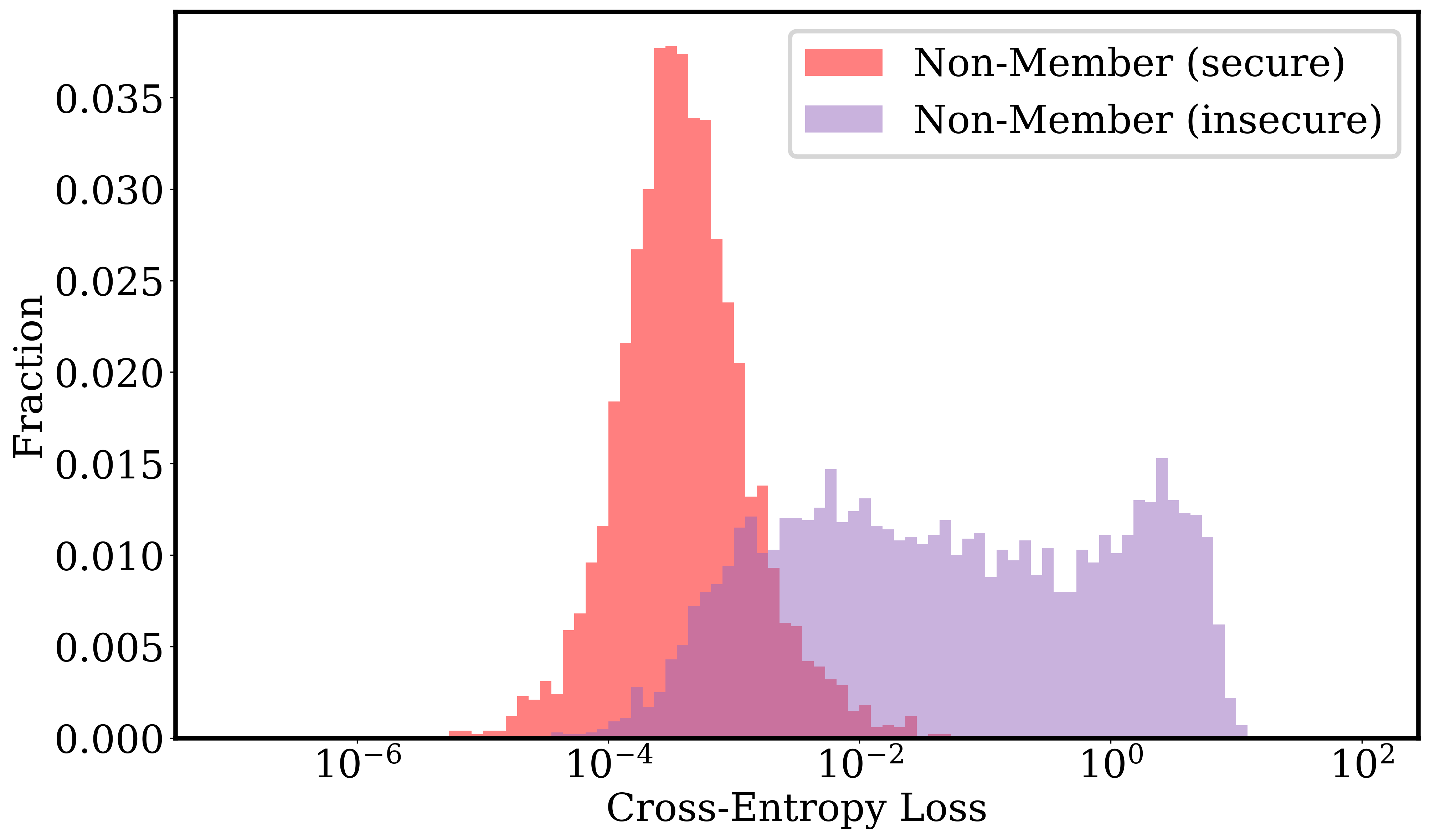}
	\caption{Histogram of the robust CIFAR10 classifier \cite{madry_robust_ICLR18} prediction loss values of both secure and insecure test examples.
	An example is called ``secure'' when it is correctly classified by the model for all adversarial perturbations within the constraint $\mathcal{B}_{\epsilon}$. }
	\label{fig:loss_vs_robustness}
\end{figure}

Note that only a few training points were not secure, so we focused our fine-grained analysis on the test set. 
Figure \ref{fig:loss_vs_robustness} shows that \textbf{insecure test inputs are very likely to have large prediction loss (low confidence value)}. 
Our membership inference strategies directly use the confidence to determine membership, 
so the privacy risk has a strong relationship with robustness generalization, even when we purely rely on the prediction confidence of the benign unmodified input.

\section{Model Architecture} \label{append:NN_Models}

We present the detailed neural network architectures used on Yale Face, Fashion-MNIST and CIFAR10 datasets in Table \ref{tab:NN_models}.

\begin{table}[!htb]
\caption{Model achitectures used on Yale Face, Fashion-MNIST and CIFAR10 datasets. ``Conv $c \, \, w \times h + s $'' represents
a 2D convolution layer with $c$ output channels, kernel size of $w \times h$, and a stride of $s$, ``Res $c$-$n$'' corresponds 
to $n$ residual units \cite{he_ResNet_CVPR16} with $c$ output channels, and ``FC $n$'' is a fully connect layer with $n$ neurons.
All layers except the last FC layer are followed by ReLU activations, and the final prediction is obtained by applying the softmax function on last FC layer.
}
\centering
\renewcommand\arraystretch{1.3}
\fontsize{6.7pt}{6.7pt}\selectfont
\begin{tabular}{|c|c|c|c|}
\hline
\multirow{2}{*}{\textbf{Yale Face}} & \multirow{2}{*}{\textbf{Fashion-MNIST}} & \multirow{2}{*}{\textbf{CIFAR10}} \\
 & & \\
\hline
Conv  $ 8 \, \, 3 \times 3 + 1$ & Conv  $256 \, \, 3 \times 3 + 1$ & Conv  $16 \, \, 3 \times 3 + 1$\\

Conv  $ 8 \, \, 3 \times 3 + 2$ & Conv  $256 \, \, 3 \times 3 + 1$ & Res 160-5\\

Conv  $ 16 \, \, 3 \times 3 + 1$ & Conv  $256 \, \, 3 \times 3 + 2$ & Res 320-5\\

Conv  $ 16 \, \, 3 \times 3 + 2$ & Conv  $512 \, \, 3 \times 3 + 1$ & Res 640-5 \\

Conv  $ 32 \, \, 3 \times 3 + 1$ & Conv  $512 \, \, 3 \times 3 + 1$ & FC 10\\

Conv  $ 32 \, \, 3 \times 3 + 2$ & Conv  $512 \, \, 3 \times 3 + 2$ &  \\

Conv  $ 64 \, \, 3 \times 3 + 1$ & FC 200 & \\

Conv  $ 64 \, \, 3 \times 3 + 2$ & FC 10 & \\

FC 200 &  & \\

FC 38 & & \\
\hline
\end{tabular}
\label{tab:NN_models}
\end{table}

\section{Experiment Modifications for the Duality-Based Verifiable Defense} \label{sec:append_changes}
When dealing with the duality-based verifiable defense method \cite{wong_robust_ICML18, wong_robust_NIPS18} (implemented in PyTorch), 
we find that the convolution with a kernel size $3 \times 3$ and a stride of $2$ as described in Section \ref{sec:setup} is not applicable.
The defense method works by backpropagating the neural network to express the dual problem, 
while the convolution with a kernel size $3 \times 3$ and a stride of $2$ prohibits their backpropagation analysis 
as the computation of output size is not divisible by 2 (PyTorch uses a round down operation).
Instead, we choose the convolution with a kernel size $4 \times 4$ and a stride of $2$ for the duality-based verifiable defense method \cite{wong_robust_ICML18, wong_robust_NIPS18}.

For the same reason, we also need to change the dimension of the Yale Face input to be $192 \times 192$ by adding zero paddings.
In our experiments, we have validated that the natural models trained with the above modifications have similar accuracy 
and privacy performance as the natural models without modifications reported in Table \ref{tab:verifiable_defenses_privacy_YALEFACE} 
and Table \ref{tab:verifiable_defenses_privacy_FMNIST}.

\section{Membership Inference Attacks with Varying Perturbation Constraints} \label{sec:different_eps_fmnist&cifar}

This section augments Section \ref{subsec:different-lp-balls} to evaluate the success of membership inference attacks when the adversary does not know the $l_{\infty}$ perturbation constraints of robust models.

\begin{figure}[!ht]
	\centering
	\includegraphics[width=\linewidth]{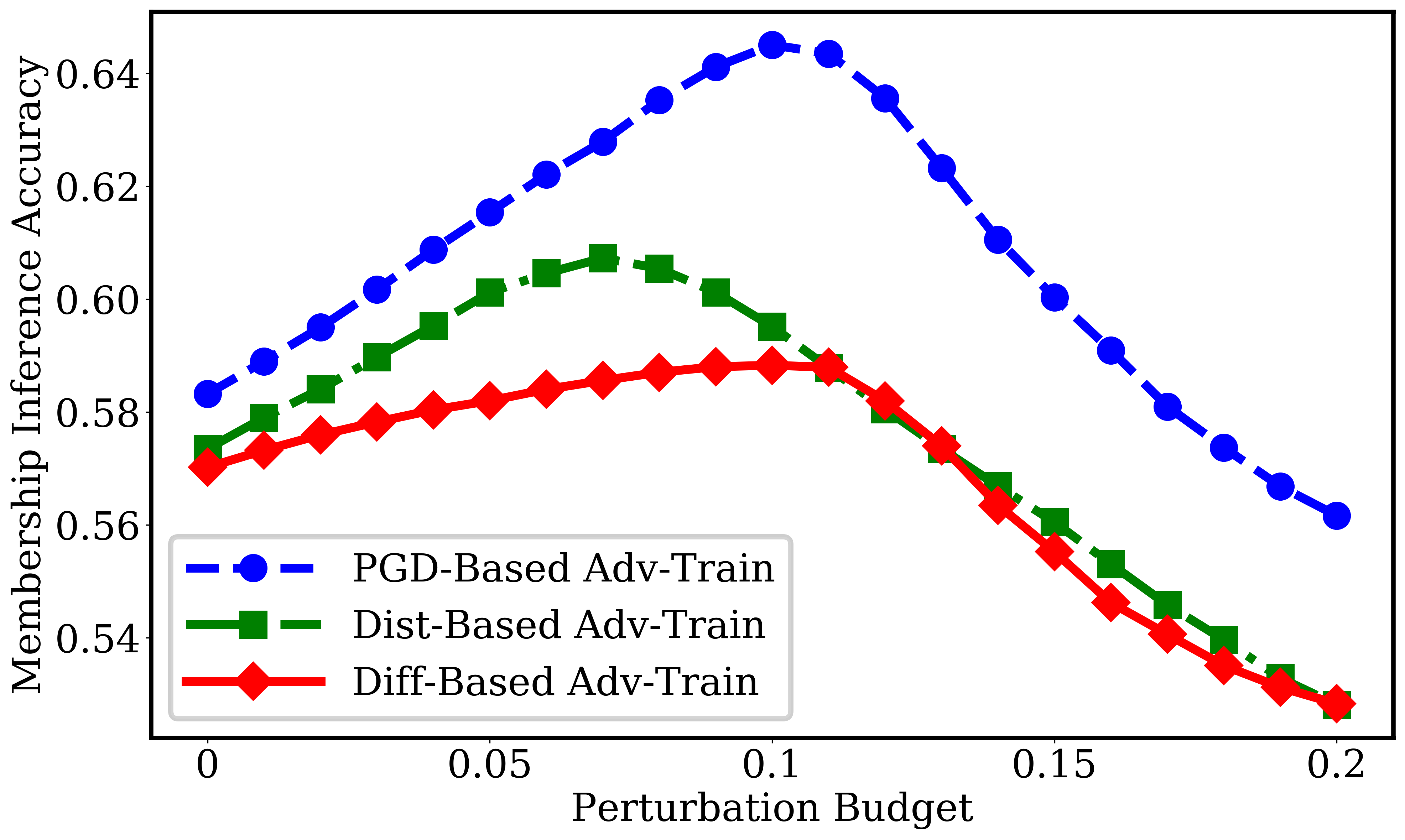}
	\caption{Membership inference accuracy on robust Fashion-MNIST classifiers \cite{madry_robust_ICLR18, sinha_PAT_ICLR18, zhang_TRADES_ICML19} trained with the $l_{\infty}$ perturbation constraint of $0.1$.
	The privacy leakage is evaluated via the inference strategy $\mathcal{I}_{\boldsymbol{\mathrm{A}}}$ based on adversarial examples generated with varying perturbation budgets.
	}
	\label{fig:fmnist_varied_eps}
\end{figure}

\begin{figure}[!ht]
	\centering
	\includegraphics[width=\linewidth]{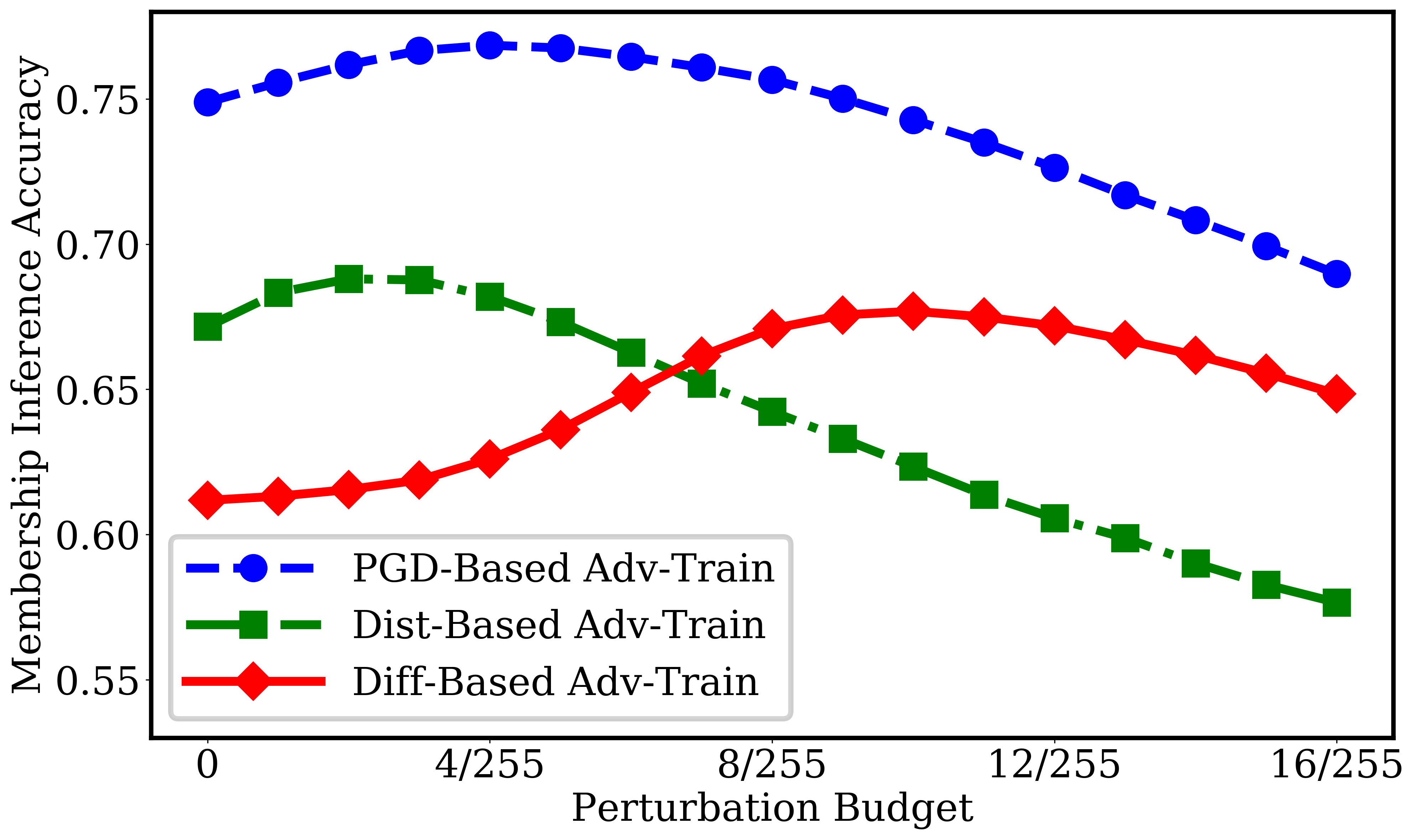}
	\caption{Membership inference accuracy on robust CIFAR10 classifiers \cite{madry_robust_ICLR18, sinha_PAT_ICLR18, zhang_TRADES_ICML19} trained with the $l_{\infty}$ perturbation constraint of $8/255$.
	The privacy leakage is evaluated via the inference strategy $\mathcal{I}_{\boldsymbol{\mathrm{A}}}$ based on adversarial examples generated with varying perturbation budgets.
	}
	\label{fig:cifar_varied_eps}
\end{figure}

We perform membership inference attacks with varying perturbation budgets on robust Fashion-MNIST and CIFAR10 classifiers \cite{madry_robust_ICLR18, sinha_PAT_ICLR18, zhang_TRADES_ICML19}.
The Fashion-MNIST classifiers are robustly trained with the $l_{\infty}$ perturbation constraint of $0.1$, while the CIFAR10 classifiers are robustly trained with the $l_{\infty}$ perturbation constraint of $8/255$.
The membership inference attack results with varying perturbation constraints are shown in Figure \ref{fig:fmnist_varied_eps} and Figure \ref{fig:cifar_varied_eps}.

\section{Privacy Risks of Other Robust Training Algorithms}\label{sec:other_robust_training}
Several recent papers \cite{lecuyer_PixelDP_SP19, cohen2019certified} propose to add a noise layer into the model for adversarial robustness. Here we evaluate privacy risks of the robust training algorithm proposed by Lecuyer et al. \cite{lecuyer_PixelDP_SP19}, which is built on the connection between differential privacy and model robustness. 
Specifically, Lecuyer et al. \cite{lecuyer_PixelDP_SP19} add a noise layer with a Laplace or Gaussian distribution into the model architecture, such that small changes in the input image with a $l_{p}$ perturbation constraint can only lead to bounded changes in neural network outputs after the noise layer.
We exploit benign examples' predictions to perform membership inference attacks ($\mathcal{I}_{\boldsymbol{\mathrm{B}}}$) against the robust CIFAR10 classifier provided by Lecuyer et al. \cite{lecuyer_PixelDP_SP19}\footnote{\url{https://github.com/columbia/pixeldp}}, which is robustly trained for a $l_{2}$ perturbation budget of $0.1$ with a Gaussian noise layer.
Our results show that the robust classifier has a membership inference accuracy of $64.43\%$. In contrast, the membership inference accuracy of the natural classifier is $55.85\%$.

\end{document}